\documentclass{iosart2c}
\usepackage[T1]{fontenc}
\usepackage{times,graphics}%
\usepackage[table,x11names]{xcolor}
\usepackage{natbib}
\usepackage{lipsum}
\usepackage{pgfplots}
\usepackage{tikz}
\usetikzlibrary{patterns}
\usepackage{ctable}
\usepackage{dcolumn}
\usepackage{graphics}
\usepackage{times}
\usepackage{helvet}
\usepackage{hhline}
\usepackage{courier}
\usepackage{makeidx}
\usepackage{amssymb,latexsym,amsmath}    
\usepackage{pgfplots}
\usepackage{ctable}
\usepackage{graphicx,array}
\usepackage{enumitem}
\usepackage{booktabs}
\usepackage{multirow} 
\usepackage{multicol} 
\usepackage{soul}
\usepackage{rotating}
\usepackage{float}
\usepackage{algorithm}
\usepackage{algpseudocode}
\usepackage{amssymb,latexsym,amsmath}     
\usepackage{graphicx,array}
\usepackage{enumitem}
\usepackage{booktabs}
\usepackage{multirow} 
\usepackage{multicol} 
\usepackage{float}
\usepackage{amssymb}
\usepackage{textcomp,algorithmicx}
\usepackage[autostyle]{csquotes}
\usepackage{commath}
\usepackage{hhline}
\usepackage{soul}
\usepackage{rotating}
\usetikzlibrary{arrows}
\usetikzlibrary{shapes}

\def\reftable#1{\mbox{Table\,\ref{#1}}}

\def\ss{\nobreak\hspace{.16667em plus .08333em}}
\newcolumntype{d}[1]{D{.}{.}{#1}}

\firstpage{1} \lastpage{5} \volume{1} \pubyear{2015}
\def\shiq{$\mathcal{SHIQ}$}
\def\bfshiq{${\bf\mathcal{SHIQ}}$}
\def\o{$\mathcal{O}$}
\def\od{$\mathcal{O'}$}
\def\L{$\mathcal{D}_\mathcal{O}$}
\def\R{$\mathcal{R}$}

\def\IM{\Im}

\newtheorem{definition}{Definition}

\def\curly#1{\{\,#1\,\}}
\def\rcurly#1{#1\,\}}
\def\labelset#1{\L(#1)}

\def\shiq{$\mathcal{SHIQ}$}
\def\o{$\mathcal{O}$}
\def\od{$\mathcal{O'}$}
\def\L{$\mathcal{L}_{\mathcal{O}}$}

\def\R{$\mathcal{R}$}

\makeatletter
\def\BState{\State\hskip-\ALG@thistlm}
\makeatother

\makeatletter
\newcommand\xleftrightarrow[2][]{%
  \ext@arrow 9999{\longleftrightarrowfill@}{#1}{#2}}
\newcommand\longleftrightarrowfill@{%
  \arrowfill@\leftarrow\relbar\rightarrow}
\makeatother

\begin{document}
\begin{frontmatter}                           

\title{Ontology Verbalization using Semantic-Refinement }
\runningtitle{Instructions for the preparation of a camera-ready paper in \LaTeX}

\review{Name Surname, University, Country}{Name Surname, University, Country}{Name Surname, University, Country}

\author{\fnms{Vinu} \snm{E.V}\thanks{Corresponding author. E-mail: vinuev@cse.iitm.ac.in, mvsquare1729@gmail.com}}
and
\author{\fnms{P Sreenivasa} \snm{Kumar}}

\runningauthor{F. Author et al.}
\address{Department of Computer Science and Engineering, Indian Institute of Technology Madras, Chennai, India\\
E-mail: \{vinuev,psk\}@cse.iitm.ac.in }
\begin{abstract}
In this paper, we propose a rule-based technique to generate redundancy-free natural language (NL) descriptions of  Web Ontology Language (OWL) entities. The existing approaches which address the problem of verbalizing OWL ontologies generate NL text segments which are close to their counterpart statements in the ontology. Some of these approaches  also perform grouping and aggregating of these NL text segments to generate a more fluent and comprehensive form of the content. 
Restricting our attention to description of individuals and atomic concepts, we find that the approach currently followed in the available tools is that of determining the set of all logical conditions that are satisfied by the given individual/concept name and translate these conditions verbatim into corresponding NL descriptions. Human-understandability  of such descriptions is affected by the presence of repetitions and redundancies, as they have high fidelity to the OWL representation of the entities. 
In the literature, no efforts had been taken to remove redundancies and repetitions at the logical level before generating the NL descriptions of entities and we find this to be the main reason for lack of readability of the generated text. In this paper, we propose a technique called \emph{semantic-refinement} to generate meaningful and easily-understandable (what we call redundancy-free) text descriptions of individuals and concepts of a given OWL ontology. 
We identify the combinations of OWL/DL constructs that lead to repetitive/redundant descriptions and  propose a series of refinement rules to rewrite the conditions that are satisfied by an individual/concept in a meaning-preserving manner. The reduced set of conditions are then employed for generating textual descriptions. 
Our experiments show that, semantic-refinement technique leads to significantly improved descriptions of ontology entities. 
We also test the effectiveness and usefulness of the the generated descriptions for the purpose of validating the ontologies and find that the proposed technique is indeed helpful in the context. 
The details of an empirical study to support the claim are provided in the paper.

\end{abstract}


\end{frontmatter}

\renewcommand\vec[1]{\overrightarrow{#1}}
\newcommand\cev[1]{\overleftarrow{#1}}
\newtheorem{example}{Example}
\def\ss{\nobreak\hspace{.16667em plus .08333em}}

\section{Introduction}
Web Ontology Language (OWL/DL) ontologies are knowledge representation structures which are based on decidable fragments of first order logic. They model domain knowledge in the form of logical axioms; so that, an intelligent agent with the help of a reasoning system, can make use of them for several applications.  Ontologies  play an important role in  the  development  and  deployment  of  the  Semantic Web  since they help in  enhancing  the understanding  of  the  contextual  meaning  of  data. Since the knowledge in the form of an ontology is inherently characterized by complex relational contexts, it is typically inaccessible for non-Semantic Web experts. This problem motivated researchers to work on natural language (NL) verbalization techniques for OWL ontologies.  The existing approaches in this direction mainly strive for one-to-one conversion of logical statements to NL texts, and result in methods which produce verbatim equivalents of OWL constructs. One of the main and common drawback of these approaches is that, since the generated sentences are verbatim equivalent to the OWL statements, they are likely to have high amount of redundancy. As we show later with examples, it can be very annoying for a human reader  to read and understand such sentences. Therefore, in this paper, we explore techniques which can generate NL sentences that do not have redundancies and are semantically equivalent to their OWL counterparts.

We will closely look at the problem of verbalization of OWL ontologies  from the perspective of using the generated descriptions for validating the formalized knowledge. 
Typically, ontologies are developed by a group of knowledge engineers with the help of domain experts. The domain experts provide the knowledge to be formalized and the engineers build the ontology out of it. Since an ontology development involves multiple parties (engineers and domain experts), the process usually follows a Spiral model, where suitable feedback mechanisms are involved to improve the structure.  

As an ontology evolves over a period of time, it can grow in size and complexity. Unless the updates are carefully carried out, the quality of the ontology might degrade. 
To prevent such quality depletion, usually an ontology development cycle is accompanied by a validation phase, where both the  knowledge engineers and domain experts meet to review the content of the ontology.

In a typical validation phase, new axioms are included or existing axioms are altered or removed, to maintain the correctness of the ontology. 
The conventional method for incorporating new axioms and validating the ontology involves a validity check by domain experts. Domain experts, who do the validity check, cannot be expected to be highly knowledgeable  on formal methods and notations. For their convenience, the OWL axioms will have to be first  
converted  into corresponding NL texts.  
Ontology verbalizers and ontology authoring tools such as ACE~\cite{kaljurand2007}, NaturalOWL~\cite{Androutso} and SWAT Tools~\cite{third2011}, 
can be utilized for generating controlled natural language (CNL) descriptions of OWL statements.
Restricting our attention to description of individuals and atomic concepts, we find that the approach currently followed in the available tools is that of determining the set of all logical conditions that are satisfied by the given individual/concept name and translate these conditions verbatim into corresponding NL descriptions.
 But the verbatim fidelity of such descriptions to the underlying OWL statements, makes them a poor choice for ontology validation. This is because, the descriptions  will be confusing to a person who is not familiar with formal constructs, and it will be difficult to correctly understand the meaning from such descriptions. This issue had been previously reported in papers such as~\cite{StevensMWPT11,third2011}, where the authors tried to overcome the issue by applying operations such as grouping and aggregation  on the verbalized text. But, since the issue had been treated  at the NL text level, the opportunity for a logical-level refinement of the OWL statements to generate a more meaningful and human-understandable representation has been ignored.

 For example, consider the following logical axioms (from People \& Pets ontology\footnote{http://www.cs.man.ac.uk/$\sim$horrocks/ISWC2003/Tutorial
/people+pets.owl.rdf}) represented in the description logic (DL) notation. 
\begin{itemize}
\item[1.]\texttt{\small Cat\_Owner $\sqsubseteq$ Person $\sqcap~\exists$hasPet.Animal $\sqcap~\exists$hasPet.Cat}

 \item[2.]\texttt{\small Cat\_Owner(sam)}

 \item[3.]\texttt{\small Cat $\sqsubseteq$ Animal}
\end{itemize}
%
%

  The different variants of the CNL sentences correspond to the individual \texttt{sam}  are as follows:
\begin{itemize}
 \item\emph{A cat-owner is a person. A cat-owner has as pet an animal. A cat-owner has as pet a cat. Sam is a cat-owner. All cats are animals}. 

or (with grouping and aggregation)

 \item\emph{ A cat-owner is a person . A cat-owner is all of the following: something that has pet an animal, and something that has pet a cat; Example: sam. All cats are animals}.
\end{itemize}
  
  As can be easily seen, these descriptions have redundant information and attempting verbatim equivalence to DL constructs has resulted in this situation. The above example illustrates one type of redundancy and several more are identified in the paper later.
  
In this paper, we introduce an approach for  removing redundancies from the verbalized  definitions of 
 OWL/DL entities, and to generate the so-called \emph{redundancy-free} representations/descriptions. 
  We  propose a technique called \emph{semantic-level refinement} (or simply  \emph{semantic-refinement}) that helps in removing the redundant (portion of the) restrictions and generating a more semantically comprehensive description of the entity. From an application point of view, in this paper, we particularly focus on generating NL descriptions of \emph{individuals} and \emph{concepts} for validating ontologies which follow \shiq~description logic.  

Our proposed approach generates NL descriptions of individuals and concepts by giving importance to the semantic  conciseness of the content. 
If we revisit our  previous example,  we expect our approach to produce a text similar to: \emph{Sam: is a cat-owner having at least one cat as pet}; such that the redundant portion of the text \emph{has as pet an animal} (since it clearly follows from \emph{having at least one cat as pet}) is removed. 


This paper is arranged as follows: Section~\ref{pre} and~\ref{new} discuss the preliminaries for understanding the work and,  newly introduced terminologies in the paper respectively. In Section~\ref{pa} we elaborate an approach for generating definitions  (in the form of logical expressions) of ontology  individuals and concepts, and a rule-based method for removing redundancies from the definitions. Section~\ref{nl} explains the process that we have followed for generating NL sentences from the logical expressions.

In Section~\ref{eval}, the empirical evaluation section, we seek to validate the following two propositions using case studies.  
Firstly, logical-level removal of redundancies and repetitions can significantly improve the clarity of  the domain knowledge when expressed in a NL. Secondly, NL definitions of  individuals of an ontology can be effectively used for validating the ontology.

\begin{table*}[!ht]
\begin{minipage}[!t]{.5\linewidth}
   \centering
   \caption {The syntax and semantics of \shiq~concept types\label{tab:t1}}
   \scalebox{0.88}{\small
   \begin{tabular}{@{}l@{~~} l l@{}}
   \toprule
    Name            &   Syntax   &  Semantics \\ 
   \midrule
    atomic concept  & $A$        & $A^{\mathcal{I}}$ \\
    top concept     & $\top$     & $\Delta^{\mathcal{I}}$\\
   bottom concept & $\bot$&$\phi$ \\
   negation & $\neg C$ &$\Delta^{\mathcal{I}}\backslash C^{\mathcal{I}}$\\
   conjunction & $C \sqcap D$&$C^{\mathcal{I}} \cap D^{\mathcal{I}}$ \\
   disjunction & $C \sqcup D$ &$C^{\mathcal{I}} \cup D^{\mathcal{I}}$ \\
   existential restriction & $\exists R.C$ &$\{\,x\in \Delta^{\mathcal{I}}\mid\exists y.\langle x,y\rangle      \in R^{\mathcal{I}}\wedge y \in C^{\mathcal{I}}\,\}$\\
   universal restriction & $\forall R.C$ &$\{\,x\in \Delta^{\mathcal{I}}\mid\forall y.\langle x,y\rangle      \in R^{\mathcal{I}}\Rightarrow y \in C^{\mathcal{I}}\,\}$\\
  min cardinality & $\geq nR.C$ &$\{\,x\in \Delta^{\mathcal{I}}\mid\#\{y\mid\langle x,y \rangle\in R^{\mathcal{I}}\land y \in C^{\mathcal{I}}\}\ge n\,\}$\\
  max cardinality & $\leq mR.C$ &$\{\,x\in \Delta^{\mathcal{I}}\mid\#\{y\mid\langle x,y \rangle\in R^{\mathcal{I}}\land y \in C^{\mathcal{I}}\}\le m\,\}$\\
  \bottomrule
  \end{tabular}}
  \end{minipage}~~~~~~~~~~~~~~~~~~~~~~
  \begin{minipage}[!t]{.45\linewidth}
  \centering
  \caption {The syntax and semantics of  \shiq~ontology axioms}
  \label{tab:t2}
     \scalebox{0.88}{\small
    \begin{tabular}{ @{}l@{~~~}l@{~~}ll@{} }
   \toprule
&Name & Syntax&Semantics \\
\midrule 
  &role hierarchy & $R\sqsubseteq S$&$R^{\mathcal{I}}\subseteq S^{\mathcal{I}}$ \\
 TBox  &role transitivity & Tran($R$) &${R^\mathcal{I}} \circ {R^\mathcal{I}}\subseteq {R^\mathcal{I}} $ \\

 &concept inclusion  & $C \sqsubseteq D$ & $C^{\mathcal{I}} \subseteq D^{\mathcal{I}}$ \\
 &concept equality  & $C \equiv D$ & $C^{\mathcal{I}} = D^{\mathcal{I}}$ \\\midrule
 &concept assertion  & $C(a)$& $a^{\mathcal{I}} \in C^{\mathcal{I}}$ \\
ABox &role assertion  & $R(a,b)$ & $\langle a^{\mathcal{I}}, b^{\mathcal{I}} \rangle \in R^{\mathcal{I}}$ \\
&inequality assertion&$a\not\approx b$& $a^{\mathcal{I}}\not = b^{\mathcal{I}} $\\
\bottomrule
\end{tabular}}
    \end{minipage} 
\end{table*}

\section{Related Work}
Over the last decade, several CNLs such as Attempto Controlled English (ACE)~\cite{kaljurand2007,kaljurand:phd}, Ordnance Survey\footnote{Great Britain's national mapping agency}'s Rabbit (Rabbit)~\cite{HartDG07}, and Sydney OWL Syntax (SOS)~\cite{CreganSM07}, have been specifically designed or have been adapted for ontology language OWL. All these languages are meant to make the interactions with formal ontological statements easier and faster for users who are unfamiliar with formal notations. Unlike the other languages~\cite{effectivenl,Jarrar06multilingualverbalization,Androutso} that have been suggested to represent OWL in controlled English, these CNLs are designed to have  formal language semantics and bidirectional mapping between NL fragments and OWL constructs.
Even though these formal language semantics and bidirectional mapping are helpful in enabling a formal check that the resulting NL expressions are unambiguous, they generate a collection of unordered sentences that are difficult to comprehend. 

To use these CNLs as a means for ontology authoring and for knowledge validation purposes, appropriate organization of the verbalized text is necessary. A detailed comparison of the systems that comprehend the NL texts is given in~\cite{StevensMWPT11}. Among such systems, SWAT tools\footnote{http://mcs.open.ac.uk/nlg/SWAT/} are one of the recent and prominent tools which use standard techniques from computational linguistics to make the verbalized text more readable. They tried to give better clarity to the generated text by grouping, aggregation and elision.  The Semantic Web Authoring (SWAT) NL verbalization tools have given much importance to the 
fluency of the verbalized sentences~\cite{third2011}, rather than removing  redundancies from their logical forms, hence have deficiencies in interpreting the ontology contents.

\section{Preliminaries}\label{pre}

\subsection{\bfshiq~Ontologies}
The description logic (DL) \shiq~is based on an extension of the well-known logic $\mathcal{ALC}$~\cite{alcx}, with added support for role hierarchies, inverse roles, transitive roles, and qualifying number restrictions~\cite{alc2}.

 We assume $N_{C}$ and $N_{R}$ as countably infinite disjoint sets of \emph{atomic concepts} and \emph{atomic roles} respectively. 
A \shiq~\emph{role} is either $R\in N_{R}$ or an \emph{inverse role} $R^{-}$ with $R \in N_{R}$. To avoid considering roles such as  $(R^{-})^{-}$, we define a function Inv(.) which returns the inverse of a role: Inv($R)=R^{-}$ and Inv($R^{-})=R.$

The set of concepts in \shiq~is recursively defined using the constructors in \reftable{tab:t1}, where $A\in N_{C}, C, D$ are concepts, $R, S$ are roles, and $n,m$ are positive integers. A \shiq~based ontology  --- denoted as a pair $\mathcal{O}=(T,A)$, where $T$ denotes  terminological axioms (also known as TBox)  and $A$ represents assertional axioms (also known as ABox) ---  is a set of axioms of the type specified in \reftable{tab:t2}. A role $R$ in $\mathcal{O}$ is \emph{transitive} if Tran($R$) $\in \mathcal{O}$ or Tran($R^{-}$) $\in \mathcal{O}$. Given an $\mathcal{O}$, let $\sqsubseteq_{\mathcal{O}} $ be the smallest transitive reflexive relation between roles $R_{1}$ and $R_{2},$ such that $R_{1}\sqsubseteq R_{2}\in \mathcal{O}$ implies $R_{1}\sqsubseteq_{\mathcal{O}} R_{2}$ and $R^{-}_{1}\sqsubseteq_{\mathcal{O}} R^{-}_{2}$. For a \shiq~ontology $\mathcal{O}$, the role $S$ in every concept of the form $\ge nS.C$ and $\le m S.C$ in $\mathcal{O}$, should be \emph{simple}, that is, $R\sqsubseteq_{\mathcal{O}} S$ holds for no transitive role $R$~\cite{Baader}.

The semantics of \shiq~is defined using \emph{interpretations}. An interpretation is a pair $\mathcal{I}=(\Delta^{\mathcal{I}},.^{\mathcal{I}})$ where $\Delta^{\mathcal{I}}$ is a non-empty set called the \emph{domain} of the interpretation and $.^{\mathcal{I}}$ is the \emph{interpretation function}. The function $.^{\mathcal{I}}$ assigns a set $A^{\mathcal{I}} \subseteq \Delta^{\mathcal{I}}$ to every $A\in N_{C}$, and assigns a relation $r^{\mathcal{I}} \subseteq \Delta^{\mathcal{I}} \times  \Delta^{\mathcal{I}}$ to every $r\in N_{R}$. The interpretation of the inverse role $r^{-}$ is $(r^{-})^{\mathcal{I}}:= \{\langle x,y\rangle\mid\langle y,x\rangle \in r^{\mathcal{I}}\}$. The interpretation is extended to concepts and axioms according to the rightmost column of \reftable{tab:t1} and \reftable{tab:t2} respectively, where 
$\#X$ denotes the cardinality of the set $X$.  

We write $\mathcal{I}\models\alpha$, if the interpretation $\mathcal{I}$ satisfies the axiom $\alpha$ (or $\alpha$ is \emph{true} in $\mathcal{I}$). $\mathcal{I}$ is a \emph{model} of an ontology $\mathcal{O}$ (written $\mathcal{I}\models \mathcal{O}$) if $\mathcal{I}$ satisfies every axiom in $\mathcal{O}$. If we say $\alpha$ is entailed by $\mathcal{O}$, or $\alpha$ is a \emph{logical consequence} of $\mathcal{O}$ (written $\mathcal{O}\models\alpha$), then every model of $\mathcal{O}$ satisfies $\alpha$. A concept $C$ is \emph{subsumed} by $D$ w.r.t. $\mathcal{O}$ if $\mathcal{O} \models C \sqsubseteq D$, and $C$ is \emph{unsatisfiable} w.r.t. $\mathcal{O}$ if $\mathcal{O} \models C \sqsubseteq \bot$. \emph{Classification} is the task of computing all subsumptions $A \sqsubseteq B$ between atomic concepts such that $A,B\in N_{C}$ and $\mathcal{O} \models A \sqsubseteq B$; similarly, \emph{property classification} of \o~is the computation of all subsumptions between  properties $R \sqsubseteq S$ such that $R,S\in N_{R}$ and $\mathcal{O} \models R \sqsubseteq S$. 

\begin{table*}\caption{TBox of ACAD ontology\label{acdtbox}}
\begin{tabular}{rcp{13cm}}
\texttt{IITStudent} &
$\equiv$ &
\texttt{Student~$\sqcap~\forall$hasAdvisor.TeachingStaff~$\sqcap$  $\exists$hasAdvisor.Professor $\sqcap~\exists$enrolledIn.IITProgramme}\\

\texttt{ IIT\_MS\_Student} &
 $\equiv$&
 \texttt{IITStudent$~\sqcap~\leq 1$\,hasAdvisor.TeachingStaff}\\

\texttt{IITPhdStudent}&
$\equiv$ &
\texttt{IITStudent $\sqcap~\geq 2$\,hasAdvisor.TeachingStaff ~$\sqcap$ $\leq\,1$\,hasAdvisor.Professor}\\

 \texttt{Professor} &
 $\sqsubseteq$ &
 \texttt{TeachingStaff}\\

 \texttt{AssistantProf} &
 $\sqsubseteq$ &
 \texttt{TeachingStaff}\\
 
$\bot$&
$\equiv$&
\texttt{Professor $\sqcap$ AssistantProf} \\

$\bot$&
$\equiv$&
\texttt{IIT\_MS\_Student $\sqcap$ IITPhdStudent} \\

\end{tabular}
\end{table*}


\begin{table}\caption{ABox of ACAD ontology\label{acdabox}}
\begin{tabular}{lc}
\texttt{IITStudent(tom)}&~~~~~~~~~~~~~~\\
\texttt{IIT\_MS\_Student(tom)}&\\
\texttt{hasAdvisor(tom, bob)}&\\
\texttt{IITPhdStudent(sam)}&\\
\texttt{hasAdvisor(sam, alice)}&\\
\texttt{hasAdvisor(sam, roy)}&\\
\texttt{AssistantProf(alice)}&\\
\end{tabular}
\end{table}

\subsection{Running Example}
In this section we introduce an example ontology (called  the academic (ACAD) ontology) which we follow throughout this chapter. We have formalized various concepts in academic domain in this ontology. The ontology is rather small, but  serves the purpose well. The TBox and ABox of the ontology is given in Table~\ref{acdtbox} and~\ref{acdabox} respectively.

\section{Newly Introduced Terminologies and Definitions\label{new}}

In this section, we introduce the terminologies and definitions  by considering ontologies whose expressivity is bound to $\mathcal{SHIQ}$ description logic. 

In this paper,  we use the words \enquote{reduction}
and \enquote{refinement} interchangeably. 
In the current context, 
\lq{Description}\rq of an ontology entity refers to its  domain-specific NL definition  generated from the ontology.

\subsection{Label-sets\label{els}}

To generate descriptions 
 of individuals in an ontology, 
we associate with each individual a set of constraints it satisfies. We call these sets as \emph{label-sets} in general. A \emph{Label-set} of an individual  is called a \emph{node-label-set}  and a \emph{label-set} of a pair of individuals  is called an \emph{edge-label-set}. The rationale behind generating these label-sets is that, since all the constraints satisfied by an individual are captured at one place, it can easily be looked up for redundancies. 

\paragraph{Node-label-set}The node-label-set of an individual is the set which contains \emph{all} the class expressions and (existential, universal and cardinality) restrictions satisfied by that individual.

\begin{definition} \label{defnls}The node-label-set of an individual $x$ (represented as \L($x$)) is defined as: 
\vspace{.2cm}
\\\vspace{.2cm}
$  \mathcal{L}_{\mathcal{O}}(x) = \{ c_{i}~|~\mathcal{O}\models c_{i}(x) \}$ \\where $c_{i} $ is of the following form: 

{\bf$c_{i} = A~|~\exists R.C~|~\forall R.C~|~ \leq nR.C~|~ \geq nR.C$}\\
Here,  $A$ is an atomic concept, $C$ is a class expression and $R$ is a role name in ontology \o, and $m$ and $n$ are positive integers. $C$ is of the following form:\\
$C = A~|~C_{1}\sqcap C_{2}~|~C_{1}\sqcup C_{2}~|~\exists R.C_{1}~|~\forall R.C_{1}~| \\~~~~~~~~~~~~~~~~~~~~~~~~~~~~~~~~~~~~~~~\leq\,nR.C_{1}~|~ \geq nR.C_{1}$, \\
where $C_{1}$ and $C_{2}$ are also class expressions.
\end{definition}

Note that, in the above definition, the first-level  expressions (the $c_{i}s$) are free from disjunctions.  If an individual satisfies a disjunctive clause (a set of independent expressions combined using disjunctions), satisfiability of each of these independent expressions can be checked and if found s included as conjunctions in the label-set. 
 Clearly, the conjunction of all the elements in the label-set of an individual will be entailed by the ontology. That is, 
$\mathcal{O}\models \big(\sqcap_{i=1}^{n} c_{i}\big) (x)$ 

An example of the node-label-set of the individual $x=$ \texttt{tom} from ACAD ontology is:
{\small
\noindent\L($x$) = \{ \,\texttt{Student,\,IITStudent,\,IIT\_MS\_Student,\\$\exists$\,enrolledIn.IITProgramme,}  \texttt{\small$\leq1$\,hasAdvisor.\\ TeachingStaff, $\forall$\,hasAdvisor.TeachingStaff,}\\ \texttt{ $\exists$\,hasAdvisor.Professor}\,\} }


\paragraph{Edge-label-set}The label-set of a pair of individuals ($x,y$) is the set that contains \emph{all}  the property relationships (role names) from the first individual to the second individual. It is represented as \L($x,y$).
\begin{definition} \label{defnls}
 \L($x,y$) is formally defined as (where $N_{R}$ is the set of all atomic roles in ontology \o):
$\text{\L}(x,y)=\{ R~|~R \in N_{R} \land \text{\o} \models R(x,y)\}.  $ 
\end{definition}

 From ACAD ontology, the edge-label-set of the pair (\texttt{tom,bob}) can be written as: \L(\texttt{tom,bob}) = \{\ \texttt{hasAdvisor}\ \}.

Although various approaches can be considered for generating label-sets, the practical method that we have adopted for generating the label-sets is explained in the next subsection.
\subsection{Label-set generation technique}
\paragraph{Node-label-set generation.} The naive method to find the node-label-set of an individual is by doing satisfiability check for all combinations of roles, concepts and restrictions types; and include them if they are true. Since, this is not a practically adoptable method for large ontologies, we generate the label-set of an individual $x$ from an ontology \o~as follows. 

Firstly, we create the corresponding \emph{inferred ontology} \od (using a reasoner).
From \od,~we find all the concept names and (existential, universal and cardinality) restrictions satisfied by the individual as follows:

\emph{Step 1:} All the concept names which are satisfied by $x$ are obtained by a simple SPARQL query. We can call it as the \emph{seed} label-set.
For example, the set of concept names, which we obtained from \od, corresponding to the individual \texttt{tom} is \{\,\texttt{\small Student,\,IITStudent, IIT\_MS\_Student}\,\}.

\emph{ Step 2:} In order to get the restrictions satisfied by $x$, we access the class definitions and class subsumption axioms corresponding to the concepts which are obtained in the first step, and then consider the existential, universal and cardinality restrictions on the right hand side of those axioms to enrich the label-set.

The right hand side of the axioms in their conjunctive normal form (CNF) is used for enriching the label-set. That is, the R.H.S. will be of the form: $c_{1}\sqcap c_{2}\sqcap (c_{3}\sqcup c_{4}\sqcup c_{5}\sqcup...\sqcup c_{k})\sqcap c_{k+1}\sqcap...\sqcap c_{k+n}$. Those clauses in the CNF which do not contain any disjunction, for examples as in $c_{1}, c_{2}$ etc. are  directly included in the label-set. If a clause contains  disjunction of expressions (denoted as D-Clause), such as $c_{3}\sqcup c_{4}\sqcup c_{5}\sqcup...\sqcup c_{k}$ above, then it is handled in parts, as shown in Algorithm~\ref{euclid}.

\begin{algorithm}
\caption{Handling disjunctions of expression}\label{euclid}
\begin{algorithmic}[1]
\Procedure{Label-set-Gen}{$x$, D-Clause}
\For{{\bf each} expression \emph{exp} in D-Clause}
   
   \If {\textit{exp} is of the form $\exists R.C$}
      \If {$\mathcal{O}\models \exists R.C(x)$}
       \State $\mathcal{L_{O}}(x)\gets\mathcal{L_{O}}(x)\cup \{\exists R.C\} $ 
      \EndIf

   
\ElsIf{\textit{exp} is of the form $\forall R.C$}
 \If {$\mathcal{O}\models \forall R.C(x)$}
       \State $\mathcal{L_{O}}(x)\gets\mathcal{L_{O}}(x)\cup \{\forall R.C\} $ 
      \EndIf

   \ElsIf {\textit{exp} is of the form $\le n R.C$}
  \If {$\mathcal{O}\models \le nR.C(x)$}
       \State $\mathcal{L_{O}}(x)\gets\mathcal{L_{O}}(x)\cup \{\le n R.C\} $ 
      \EndIf

    \ElsIf {\textit{exp} is of the form $\ge n R.C$}
  \If {$\mathcal{O}\models \ge nR.C(x)$}
       \State $\mathcal{L_{O}}(x)\gets\mathcal{L_{O}}(x)\cup \{\ge n R.C\} $ 
      \EndIf
   \EndIf
   
\EndFor

\EndProcedure
\end{algorithmic}
\end{algorithm}

\def\Lof#1{\L($#1$)}
\def\Rof#1{\R\big(#1\big)}
\def\setof#1{\{#1\}}
\def\Foreach{\For\kw{each}\ }
\def\Forall{\For\kw{all}\ }
%
%
%
%
%
%
%

Continuing with our example, enrichment of the label-set of \texttt{tom} is done by obtaining  existential, universal and cardinality restrictions associated with each of the concept names in the seed label-set. That is, the restrictions  \texttt{\small $\exists$\,enrolledIn.IITProgramme,\\ $\leq1\,$hasAdvisor.TeachingStaff, }\\ \texttt{\small$\forall$\,hasAdvisor.TeachingStaff,} and \texttt{\small\\ $\exists$\,hasAdvisor.Professor} associated with concept names are included in \L(\texttt{tom}).

 It should be noted that, using this approach, we are generating  
 only those necessary restrictions which can entail the other satisfying combinations as per our  label-set definition. For the same reason, we may need to rely on rule-based reasoning (explained later) to generate other restrictions which are of our interest.

\newcolumntype{P}[1]{>{\raggedright\arraybackslash}p{#1}}
\def\curly#1{\{\,#1\,\}}
\def\rcurly#1{#1\,\}}
\def\labelset#1{\L(#1)}
\begin{table*}[ht]
\vspace{.5cm}
\centering \caption {Node-Label-set of individuals in ACAD ontology (intentionally omitted $\top$ class from the label-sets)\label{ls}}
\scalebox{1}{
{\renewcommand{\arraystretch}{1.25}
\begin{tabularx}{\textwidth}{ @{}l @{\;}>{=~\{}c@{~} X }
\toprule
\L (\texttt{tom}) & & \rcurly{\texttt{Student, IITStudent, IIT\_MS\_Student, $\exists$\,enrolledIn.IITProgramme,~~~   ~~~$~~~~~$ $\leq 1$\,hasAdvisor.TeachingStaff, $\forall$\,hasAdvisor.TeachingStaff, $\exists$\,hasAdvisor.Professor }}\\

\L(\texttt{sam}) & & \rcurly{\texttt{Student, IITStudent, IITPhdStudent,   $\exists$isEnrolledIn.IITProgramme,~~~~~$~~~~~~~$
$\geq2\,$hasAdvisor.TeachingStaff, $\leq\!\!\!1$\,hasAdvisor.Professor, $\forall$hasAdvisor.TeachingStaff,
$\exists$hasAdvisor.Professor}}
\\

\L(\texttt{bob}) & & \rcurly{\texttt{Professor, TeachingStaff}}
\\


\L(\texttt{alice}) & & \rcurly{\texttt{AssistantProf, TeachingStaff}}
\\



\L(\texttt{roy}) & & \rcurly{\texttt{Professor, TeachingStaff}}
\\
\toprule
\end{tabularx}}}
\end {table*}



\paragraph{Edge-label-set Generation} The edge-label-set of a pair of individuals ($x,y$)  can be easily generated from \od~using a simple SPARQL query.

\section{Proposed Method for Generating Descriptions}\label{pa}

Once we get the label-sets of all the  individuals (node-label-sets) 
in a given ontology, we can generate  descriptions of individuals and concepts using the following approaches. 

\subsection{Description of individuals}

Node-label-sets of each individuals are considered for generating their descriptions. Label-sets of all the individuals from ACAD ontology is given in Table~\ref{ls}.
For example, by looking at the node-label-set of \texttt{tom}, we will get  the set of all restrictions (logical expressions) that are satisfied by the individual. Considering  these restrictions together, we can frame a meaningful definition for \texttt{tom} as: \emph{\enquote{Tom is a student who is enrolled in an IIT Programme,  has one professor as advisor, and all his advisors  are teaching staffs.}} Clearly, not all logical expressions (labels) in the label-set are necessary to generate such a description.  That is, those labels that can induce redundancy in the description can be ignored or combined with other restrictions.

As noted earlier, some of the labels (mainly role restrictions) in the label-set if verbalized directly may generate confusing descriptions, and hence they should be reduced or combined with other restrictions to get a more refined restriction. For  example, if left unrefined, the restrictions \texttt{\small$\forall$hasAdvisor.TeachingStaff} and \texttt{\small$\forall$hasAdvisor.$\top$}   may give rise to the description: \emph{\enquote{all advisors are some one and all advisors are teaching staffs}}, which  confuses a human reader.

Given a label-set, the naive method to remove redundant labels is by considering combinations of labels and trying to see whether they can be reduced or not. This is indeed a tedious process, since the total number of steps to be taken for complete reduction depends of the combination which we select at each step. To overcome this, we propose a rule-based process where labels of a specific restriction types are handled in a pre-defined order. A systematic method utilizing a set of rules which will always generate stricter (more specific) forms of a given set of restriction, is also proposed to  attain complete refinement of the label-sets. Due to the aforementioned property of the rules, we call them as \emph{refinement-rules}.  Since we do this reduction or refinement of labels  at the logical-level by considering their semantics, we call this rule-based refinement process as \emph{semantic-refinement of label-sets}. The refined form of the label-set is called \emph{semantically-refined label-set}.


The semantic-refinement is not only done to remove redundant labels in a label-set, but also to avoid ambiguous verbalization of interim logical expressions. For example, \texttt{\small $\forall$hasAdvisor.Professor} is a label which can appear in the label-set of an individual of  \texttt{\small IITStudent} due to the axiom:  \texttt{\small IITStudent  $\sqsubseteq$ $\forall$hasAdvisor.Professor}.  Linguistically this label (along with the axiom) can be interpreted in two ways. That is, either as \emph{All advisors of IIT students are teaching staffs} or, by considering logical equivalent of the statement, it can be interpreted as \emph{Either all advisors of IIT students are teaching staffs or} (vacuously-true case) \emph{they do not have an advisor.} Clearly, including the latter description in the verbalization may confuse a reader. This is especially the case when it can be inferred from other axioms that vacuously-true case does not arise. 

For identifying the cases where combinations of conditions involving qualifiers and/or number restrictions occur and to succinctly represent them, we introduce the following new constructors.

\begin{itemize}
\item {Non-vacuous} role restriction: {$\Im R.C$}\\
$\Im R.C^{\mathcal{I}}= \{ x\in \Delta^{\mathcal{I}}|\exists y. \langle x,y\rangle \in {R^\mathcal{I}}\land y \in C^{\mathcal{I}}\land$ $\forall z. \langle x,z\rangle \in {R^\mathcal{I}} \implies z \in C^\mathcal{I} \}$

\item {Exactly-one} role restriction: {$\exists_{=1}R.C$}\\
$\exists_{=1}R.C^{\mathcal{I}}= \{ x\in \Delta^{\mathcal{I}}| (\exists y_{1}.\langle x,y_{1}\rangle \in {R^\mathcal{I}} \land y_{1} \in C^\mathcal{I} \land $ $\exists y_{2}.\langle x,y_{2}\rangle \in {R^\mathcal{I}} \land y_{2}\in C^\mathcal{I})\implies y_{1} = y_{2}\}$

\item {Exactly-$n$} role restriction: {$\exists_{=n}R.C$}, general case of exactly-one role restriction.
\end{itemize}

In our rule-based refinement process, like any rule-based approach, the order at which the rules are applied is important, as the applicability of one rule may depend on another. We observed that there is a notion of \emph{strictness} associated with role restrictions which can be effectively utilized for ordering the rules. The notion of strictness can be looked at as:  if a role restriction $R_{1}$  is implied by another role restriction $R_{2}$ (i.e., $R_{2}\implies R_{1}$), then $R_{1}$ can be said as a stricter version of $R_{2}$. For instance, $\Im R.U$ can be said as the stricter form of  $\exists R.U$ and $\forall R.U$. Similarly, $\exists_{=n}R.U$ is a sticker form of $\le nR.U$ and $\ge nR.U$. 
Since we intend to find sticker forms of role-restrictions, the obvious way is to  apply rules corresponding to less stricter restriction types  prior to those  of stricter restriction types. 

%


In the forthcoming sub-section, we introduce our rule-based refinement algorithm to accomplish complete reduction, where, we do all the possible reduction of less stricter restrictions prior to reducing stricter ones. Completeness of the refined form of label-set is guaranteed by the construction of the algorithm.



In what follows, we discuss how semantic-refinement of label-sets can be achieved.  

\subsubsection{Semantic-refinement of label-sets} 
We propose seven sets of rules for refining a label-set. Each of these rule sets contain carefully chosen rules which are repeatedly applied to the restrictions in the label-set, until no more reduction is possible. On moving from one rule set to another, those labels which have been  reduced would be  \emph{provisionally} removed from the label-set. More details about the algorithm is given in the next sub-section.

\begin{table*}[ht]\caption{Details of rule sets 1-5.\label{12345}}
\begin{tabular}{l l l l l }
\toprule
{\bf Rule No.}&{\bf Restriction 1}&{\bf Restriction 2}&{\bf Condition}&{\bf Refined form}\\\toprule
\multicolumn{5}{l}{\bf Concept Refinement rule}\\
1a&\multicolumn{4}{l}{ Concept names, whose (equality) definitions are already    }\\
&\multicolumn{4}{l}{included in the label-set, can be removed. }\\
\midrule

\multicolumn{5}{l}{\bf Superclass  Refinement rule}\\

2a&$U$ & $V$ &$U \sqsubseteq V $ &$U$ \\
\midrule

\multicolumn{5}{l}{\bf Existential Role  Refinement rule}\\
3a&$\exists R.U$ & $\exists S.V$ &$U \sqsubseteq V ~\& ~\ R \sqsubseteq S$ &$\exists R.U$ \\\midrule

\multicolumn{5}{l}{\bf Universal Role  Refinement rules}\\
4a&$\forall R.U$ & $\forall S.V$ &$U \sqsubseteq V ~\& ~\ S \sqsubseteq R$ & $\forall R.U$, $\forall S.U$ \\
   
4b& $\forall R.U$ & $\forall R.V$ &$V \sqsubseteq U$ & $\forall R.V$ \\\midrule
     

\multicolumn{5}{l}{\bf III \& IV Combination rules}\\   
5a& $\exists R.U$ & $\forall R.U$ &  &$\Im R.U$  \\
5b&  $\forall R.U$ & $\exists S.V$ &$U \sqsubseteq V ~\& ~\ S \sqsubseteq R$ &$\Im R.U$, $\Im S.U$ \\  

5c&  $\forall R.U$ & $\exists S.V$ &$V \sqsubseteq U ~\& ~\ S \sqsubseteq R$ &$\Im R.U$, $\exists S.V$ \\\bottomrule    


\end{tabular}
\end{table*}

The details of the first five sets of rules are given in Table~\ref{12345}. Each of the rule sets are given names that  correspond to the type of restriction they handle. For example, the first rule set is called \emph{Concept Refinement rule}, where atomic concepts  in the label-set are looked at for refinement. More details about the refinement rules are given below.

\paragraph{Concept Refinement Rule.} Here, we consider all the concept name symbols that are present in the label-sets and, check whether their definitions (i.e., the set of restrictions which defines the concept) are included in the label-set. If the  defining restrictions of a concept are present in the label-set, the concept name can be removed, since it is a redundant content.

\paragraph{Superclass Refinement Rule.} Consider the individuals given in Table~\ref{ls}, we can see that their label-sets contain all the concept names which they belong to. Some of the concepts in these label-sets are  hierarchically related (in class - super-class relationship) in the ontology, resulting in redundant labels.
For example, consider the label-set \L(\texttt{tom}), it contains the concepts \texttt{\small IIT\_MS\_Student} and \texttt{\small IITStudent}. Since it can be inferred from the concept \texttt{\small IIT\_MS\_Student} that \texttt{\small tom} is also a \texttt{\small IITStudent}, we can say that \texttt{\small IITStudent} is a redundant information (label) in the label-set. We remove such redundant labels  by using most-specific concept notion. 
Also an individual may be present in 2 or more such subsumption concept chains.
In each chain we need to use the most-specific concept.



 (Note that, this refinement rule is applied only after the applications of the concept refinement rule -- some specialized concepts may get removed while applying the rules in the first rule set, therefore, it does not always mean that a refined label-set contains only specialized concept names)

The presence of redundant concept names in a node-label-set is mainly because,  we do a classification on the ontology prior to the label-set generation.


The upcoming rule sets are meant for reducing the various role restrictions allowed in a \shiq~ontology. 

\begin{table*}[ht]\caption{Details of rule sets 6 and 7.\label{678}}
\begin{tabular}{l@{\,~} l l p{4cm} @{~~~~}p{4cm} }
\toprule
{\bf Rule No.}&{\bf Restriction 1}&{\bf Restriction 2}&{\bf Condition}&{\bf Refined form}\\\midrule
   
\multicolumn{5}{l}{\bf Qualified Number Restriction Refinement rules}\\  
6a&$\geq n R.U$ & $\geq m S.V$ &$U \sqsubseteq V ~\& ~\ R \sqsubseteq S$ $~\& ~\ n \geq m$ &$\geq n R.U$ \\ 

6b&$\exists R.U$ & $\geq n S.V$ &$V \sqsubseteq U ~\& ~\ S \sqsubseteq R$  $~\& ~\ n \geq 1$ &$\geq n S.V$ \\

6c&$\exists R.U$ & $\leq n R.V$ &$U \sqsubseteq V~\& ~\ n = 1$ & $\exists_{=1}R.U, \exists_{=1}R.V$ \\
            

{6d}& 
{$\geq nR.U$}& 
{$\leq n S.V$} &
$R \sqsubseteq S ~\& ~U\sqsubseteq V$ &$\exists_{=n}R.U, \exists_{=n}S.V$\\\midrule

\multicolumn{5}{l}{\bf Exactly-$n$ Role  Refinement rules}\\  

7a&$\exists R.U$ & $\exists_{=1} S.V$ &$U \sqsubseteq V ~\& ~\ R \sqsubseteq S$ &$\exists_{=1} R.U, \exists_{=1} S.V$ \\



7b&$\Im R.U$ & $\exists_{=1} S.V$ &$U \sqsubseteq V ~\& ~\ R \sqsubseteq S$ &$\exists_{=1} R.U, \exists_{=1} S.V, \Im R.U$ \\


  
{7c}& {$\geq m R.V$}&{$\exists_{=n} R.U$}  & {$U \sqsubseteq V ~\&~m\geq n$ } &{$\exists_{=n} R.U, \geq (m-n) R.(V\sqcap \neg U)$}\\\bottomrule
\end{tabular}
\end{table*}

\paragraph{Existential Role Refinement rule.} According to this rule, if a label-set contains two labels of the form: $\exists R.U$ and $\exists S.U$, and if they satisfy the condition: $U \sqsubseteq V \& R\sqsubseteq S$, then they can be refined to $\exists R.U$. In general, all these rules are defined such that given a refined form and the condition which have been used for refinement, the non-refined forms of the restriction(s) can be traced back. This means that, the refinement is done without affecting the semantics/meaning of the restrictions. Formally, the correctness of the rule can be proven as follows:

\paragraph{Proof of Rule 3a.} Given an ontology \o~with $R$ and $S$ as its roles, and $U$ and $V$ are two of its concepts, and \o$\models U\sqsubseteq V, R\sqsubseteq S$, then $\exists R.U~\sqcap~\exists S.V\equiv \exists R.U$. 
To prove this, let us consider an individual $x \in \exists R.U~\sqcap~\exists S.V$, clear it implies $x \in \exists R.U$. Therefore $\exists R.U~\sqcap~\exists S.V\sqsubseteq \exists R.U$.  Now, if $x \in \exists R.U$, it implies that there exist an arbitrary $a$, such that $(x,a)\in R, a \in U$. Since $U\sqsubseteq V$, we can say that $a\in V$. It implies, $x\in \exists S.V$. Similarly, since $R\sqsubseteq S$,   $(x,a)\in R \implies  (x,a)\in S$ Therefore, $\exists R.U \sqsubseteq \exists R.U~\sqcap~\exists S.V$. 

%
%

\paragraph{Universal Role Refinement rules.} This rule set contains two rules which help in refining universal role restrictions. If a label-set contains two role restrictions of the form: $\forall R.U$ and $\forall S.V$, universal role refinement rules can be applied if they satisfy the conditions of the rule. For example, if the label-set contains \texttt{\small$\forall$hasAdvisor.Professor} and \texttt{\small$\forall$hasAdvisor. TeachingStaff}, and if  \texttt{\small Professor  $\sqsubseteq$ TeachingStaff}, we can refine those restrictions  to  \texttt{\small$\forall$hasAdvisor. Professor}. The correctness of the two rules can be easily be proven as follows. 

\paragraph{Proof of Rule 4a.} Given an ontology \o~which entails $U\sqsubseteq V$ and $S\sqsubseteq R$ (where $R$ and $S$ are roles, and $U$ and $V$ are concepts), then $\forall R.U~\sqcap~\forall S.V \equiv \forall R.U~\sqcap~\forall S.U.$ Proving  $ \forall R.U~\sqcap~\forall S.U \sqsubseteq \forall R.U~\sqcap~\forall S.V$ is trivial since $\forall S.U \sqsubseteq \forall S.V$~~(given, $U\sqsubseteq V$). Now, let $x\in \forall R.U \sqcap \forall S.V$, suppose $(x,a)\in S$ where $a$ is an arbitrary individual. Since $S\sqsubseteq R$, $(x,a)\in R$.  It implies $a\in U$ (since $x\in \forall R.U$). Therefore, we get $x\in\forall S.U.$ Hence, $\forall R.U~\sqcap~\forall S.V \sqsubseteq \forall R.U~\sqcap~\forall S.U.$

\paragraph{Proof of Rule 4b.} Given an ontology \o~which entails $V\sqsubseteq U$ (where $R$ is a role and, $U$ and $V$ are concepts), then $\forall R.U~\sqcap~\forall R.V \equiv \forall R.V$. Proving $\forall R.U~\sqcap~\forall R.V \equiv \forall R.V$ is trivial, since the L.H.S. can be written as $\forall R.(U~\sqcap~V)$, and it is equivalent to $\forall R.V,$ since   $V\sqsubseteq U$.\\

Further in this section, we refrain from giving the proof of correctness of the rules in the succeeding rule sets. An appendix is provided at the end of the paper with all the required proofs.
%

\paragraph{III \& IV Combination rules.} In this rule set, we refine the existential and universal role restrictions which are present in the label-set.\vspace{.5cm}

The details of the next set of rule sets are given in Table~\ref{678}.

\paragraph{Qualified Number Restriction Refinement rules.} In this set there are four rules. Here we mainly try to refine qualified number restriction restrictions (of the form $\leq n R.U$ or $\geq m S.V$) to stricter version of the same form or to a exactly-$n$ restrictions.



\paragraph{Exactly-$n$ Role Restriction rules.} In this rule set, we reduce the exactly-n role restrictions which are generated using the preceding rule-sets. The rule set is named so because, this is the only rule set where we try to reduce exactly-n role restrictions.


\subsubsection{Algorithm for semantic-refinement}

As we mentioned before, semantic-refinement helps in refining restrictions, which are present in a label-set, to their stricter forms by combining them using a set of rules. The rules are applied sequentially from rule-set 1 to 7. While applying the rules, on moving from one rule-set to another, provisional removal of reduced restrictions is done to reduce  computational complexity. In our algorithm, we will mark such restrictions as PRs (Provisionally Reduced ones), so that at a later stage we can remove them permanently from the label-set.

Algorithm-2 describes the steps that has to be followed for applying the rules. This algorithm works by taking pairs of restrictions from the label-set, and looking for the applicability of the rules. If a rule is applicable, the restrictions will be checked for the following set of conditions, to decide whether to resume the reduction or not. These conditions are followed mainly to ensure quick reduction.

\begin{algorithm}
\caption{Semantic-refinement of label-sets}\label{srls}
\begin{algorithmic}[1]
\Procedure{Semantic\_Refinement}{\L(x)}
\State Mark all $u 
\in$ \L$(x)$ as not PRs
\State Apply \emph{Concept Refinement rule} and remove $~~~~~~~$appropriate concept names from \L(x)
\State $R\gets$ Rule-sets 2-7 $\triangleright$ list of pre-defined rules
\For{{\bf each} rule-set $rs \in R$}
   \State Set $M, REF \gets \phi$

 \For{{\bf each} $(u,v)$ $ \in$ \L$(x) \times$ \L$(x)$ AND $~~~~~~~~~~~~~u\not = v$}

   \If {\small(NOT(MARKED\_AS\_PR($u$)) AND $~~~~~~~~~~~~~~~~~~~~~~~$NOT(MARKED\_AS\_PR($v$)))}
     
       \For{{\bf each} ($r \in rs$) }
        
      \If {$r$ is applicable on $(u,v)$}
       \State {\small$M\gets$ APPLY\_RULE$(r,u,v)$}
       \State {\small\L$(x)\gets$ \L$(x)\cup M$}
       \State {\small$REF\gets REF~\cup~$\{u,v\} }
       \If {$u\in M$}
       \State {\small$REF\leftarrow REF\backslash\{u\}$ }
        \EndIf
      \If {$v\in M$}
       \State {\small$REF\leftarrow REF\backslash\{v\}$ }
       \EndIf
      \EndIf
  
\EndFor

\EndIf

\EndFor

\State MARK\_AS\_PR($REF$)
\State \L$(x) \gets $\L$(x)~\cup~REF$

\For{{\bf each} $u \in $\L$(x)$}
\If{\small the restn. type of $u$ is not used in the suc-$~~~~~~$cessive rule-sets AND MARKED\_AS\_PR($u$)}
\State\L$(x)\gets$\L$(x)\backslash \{u\}$
\EndIf
\EndFor

   
%
%
%
%
%
%
   
\EndFor

\EndProcedure
\end{algorithmic}
\end{algorithm}

\begin{figure*}[th!]
  \centering
  \includegraphics[width=1.03\textwidth]{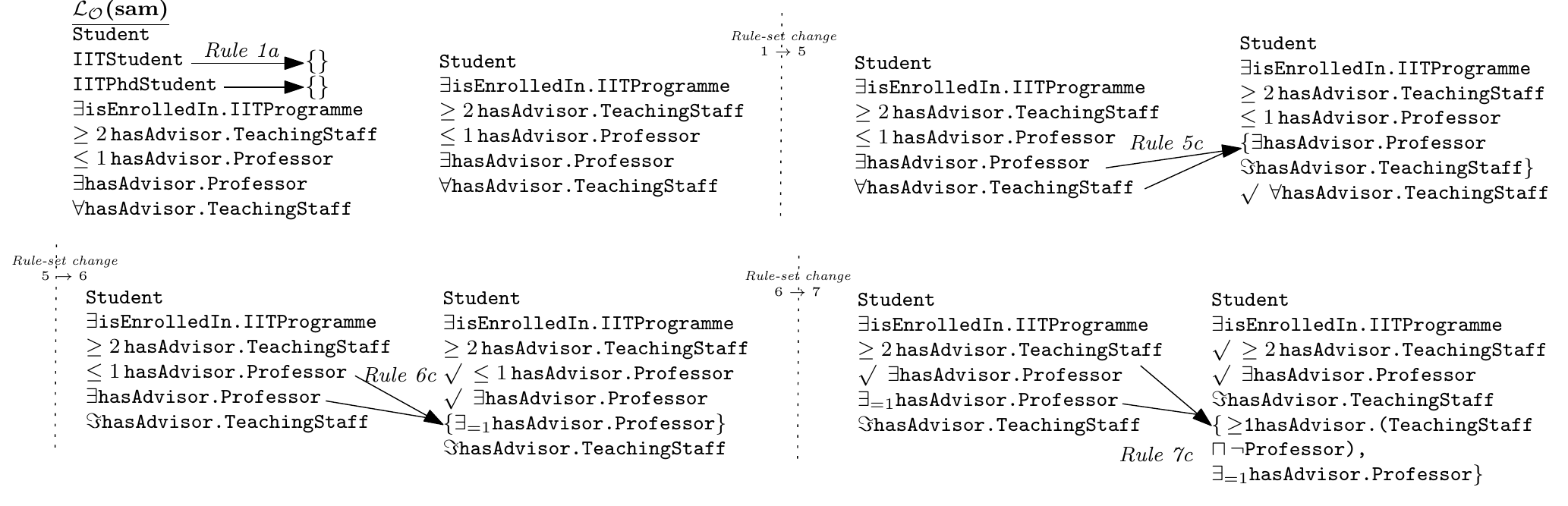}
  \caption{Steps involved in the semantic-refinement of \L(\texttt{sam}). Arrows represent the application of rules. \label{fig1}}
\end{figure*}

\paragraph{Condition-1.} No need to further reduce two provisionally reduced (PR) restrictions. (This is because, the rule-sets are designed in such a way that if a particular combination of restriction types is reduced by a rule in one rule-set, the same combination will not occur in the succeeding rule-sets)


\paragraph{Condition-2.} If a rule combines two restrictions ($R1$ and $R2$) and generates either $R1$ or $R2$, then that $R1$ or $R2$ should not be marked as a PR. ( This is because, in the rules such as 4a, 5c, 6a etc., one of their  antecedent   term gets repeated in the consequent part of the rule to ensure reverse implication (i.e., for preserving semantics). On applying such rules, if the regenerated terms are marked as PR, they may get permanently removed during the course of the algorithm, which is not acceptable.)

\paragraph{Condition-3.} If the restrictions of a particular form are  \emph{not} used in successive rule-sets, the PR restrictions of that form can be removed. (Either they can be removed after the applications of rules in all the rule-sets or they  can be removed at specific points where it can be determined that they will not be used by any rules from then on. The latter is computationally efficient)

For illustration, let us consider the node-label-set of the individual  \texttt{sam}. Fig~\ref{fig1} shows the refinement steps and the rules in the rule sets which are used for the refinement. \L(\texttt{sam}) is represented vertically. In the figure, the arrows represent the application of rules. Rule numbers are represented in italics. A refinement of two restrictions may sometimes result in more than one restrictions, to represent them, the arrows are followed by brace brackets (\{...\}) to show the resultant restrictions. 

Initially, the algorithm marks all the labels in the label-set as not PR. Then the algorithm looks for the applicability of the rule 1a (concept refinement rule). In the figure, \L(\texttt{sam}) contains the labels  \texttt{\small IITStudent} and \texttt{\small IITPhdStudent} whose definitions are present in the label-set. Therefore, the Rule 1a is applied on those labels and remove them from the label-set. 
In the algorithm, lines 5-31 take the rest of the rule-set one at a time, and look for possible application of rules on pairs of restrictions in the label-set. In our example label-set, since no rules in the rule-sets 2,3, and 4 are applicable,  we move to the rule set 5. Now, the algorithm applies the rule 5c on two of the restrictions as shown in the figure and refine them to the two restrictions given in the brackets. Application of a rule will be done only if the restrictions in the pair are not marked as PR (checked using the function MARKED\_AS\_PR(.)). The \emph{if} condition in the line-8 of the algorithm will take care of this. After the application of a rule (using the function APPLY\_RULE(.)), the details of the reduced restrictions will be stored in the set variable $REF$. Based on the condition-2, appropriate changes have to be done on the contents of $REF$ (lines 14-20). Once all the possible rules in a particular rule set are applied, the reduced restrictions will be marked as PRs (lines 24). Once the algorithm considers all pairs of labels and checks them for the applicability of all the rules in the current rule-set, the condition-3 will be checked for possible permanent removal of the PRs. The entire process will be repeated for all the succeeding rule-sets.

Coming back to our example label-set, after the application of Rule 5c, one of the  reduced restriction is marked as PR (represented using $\surd$), while the other restriction is not marked as PR due to the condition-2. On changing the rule-set, since no other rules  in rule-set 5 are applicable, the one which is marked as PR can be permanently removed since the condition-3 is satisfied. In the forthcoming iterations of the for loop (line 5), rules in the rule-set 6 and 7 are applied in similar fashion. In the last iteration, we will get the most refined 
set of labels, along with a set of restrictions which are marked as PRs. 
The restrictions which are marked as PRs are removed to get the refined label-set.

\paragraph{Illustration of the usefulness of the approach.} The usefulness of semantic refinement can be illustrated by looking at the sentences that can be generated from the node-label-set before and after refinement. Considering the original node-label-set, \texttt{sam} can be defined as \enquote{\emph{A student, an IIT student, an IIT PhD student, who is enrolled in an IIT programme, has more than two advisors who is a teaching staff, has less than one and at least one advisor who is a professor, and all advisors are teaching staff}}. By making use of  the refined node-label-set, we can generate a smaller  and easily-understandable definition: \enquote{\emph{A student who is enrolled in an IIT programme, has exactly one advisor who is a professor and has at least one more advisor who is a teaching staff but not a professor}}.  More examples and evaluation results to support the usefulness of this approach are presented in Section~\ref{eval}.

\subsection{Description of Concepts}
A concept can be defined in a similar fashion as that of an individual using label-sets. To generate the description of a concept, we introduce a new individual as its member. 
%
It is important that the new individual should be assigned  as the member of only the concept whose definition has to be found. Now, label-set corresponding to this newly introduced individual is utilized to generated the concept's definition. The rationale behind introducing a new individual is that, in order to find the definition of a concept (say Concept A), we only need restrictions which are associated with it and its super-classes. Considering an existing individual may result in a case where it may belong to concepts which are sub-classes of the concept A; this  results in including the restrictions associated with the specific-classes also in the label-set, which is undesirable. Introducing a new individual will overcome this issue; in addition, the approach will even work smoothly for those concepts which do not have an individual.

\begin{table}[!ht]\caption{\parbox[l][.8cm][c]{7cm}{Constraint-specific templates of the possible restrictions in a redundancy-free description-set.}\label{t1}}
\centering
\begin{tabular}{@{}l@{ }l@{}}\midrule
Restrictn. & Constraint-specific template\\\midrule
$\exists R.C$ & $<\!\text{R-}\texttt{verb}\!>$ at least one $<\!C\!>$ as $<\!\text{R-}\texttt{noun}\!>$\\
$\forall R.C$ & $<\!\text{R-}\texttt{verb}\!>$ only $<\!C\!>$ as $<\text{R-}\texttt{noun}>$\\
$\ge\!nR.C$ & $<\!\text{R-}\texttt{verb}\!>$ at least $<\!n\!><\!C\!>$ as $<\!\text{R-}\texttt{noun}\!>$\\
$\le\!mR.C$ & $<\!\text{R-}\texttt{verb}\!>$ at most $<\!m\!><\!C\!>$ as $<\!\text{R-}\texttt{noun}\!>$\\
$\IM R.C$ &  $<\!\text{R-}\texttt{verb}\!>$ at least one $<\!C\!>$ and\\&~~~~~~~~~~~~~~~~~~~~ only $<\!C\!>$ as $<\!\text{R-}\texttt{noun}\!>$\\

$\exists !R.C$ & $<\!\text{R-}\texttt{verb}\!>$ exactly one $<\!C\!>$ as $<\!\text{R-}\texttt{noun}\!>$\\\midrule
\end{tabular}
\end{table}

Let us look at an illustration of generating definition of \texttt{\small IITPhdStudent} from ACAD ontology. At first, we introduce the individual \texttt{\small ips} as a member of \texttt{\small IITPhdStudent}. Now we will find the label-set of \texttt{\small ips}. 

We get \L(\texttt{ips}) as {\small\{\texttt{Student, IITStudent, IITPhdStudent,   $\exists$isEnrolledIn.IITProgramme,
$\geq2$~hasAdvisor.TeachingStaff,\\ $\leq1$ hasAdvisor.Professor, $\forall$hasAdvisor.\\TeachingStaff,
$\exists$hasAdvisor.Professor} \}}

In the next step,  we remove the concept name, whose definition has to be found, from the obtained label-set. That is, {\small \L(\texttt{ips})$\backslash$\{\texttt{IITPhdStudent}\}.} This new label-set is semantically-refined and verbalized to get the redundant-free description of the concept.

\begin{table*}[!ht]\caption{Refined node-label-sets of individuals in ACAD ontology\label{rlsT}}
\begin{tabular}{lp{12cm}}\toprule
{\bf Individual}&{\bf Refined-label-set}\\
\texttt{sam} & \texttt{\{  Student, $\exists $isEnrolledIn.IITProgramme, $\exists_{=1}$hasAdvisor.Professor, $\Im$hasAdvisor.TeachingStaff,  $\geq1$hasAdvisor.(TeachingStaff $\sqcap\,\neg$Professor)\}}\\

\texttt{tom} & \texttt{\{  Student, $\exists $isEnrolledIn.IITProgramme, $\exists_{=1}$hasAdvisor.Professor \} } \\

\texttt{bob}&  \texttt{\{ Professor \} } \\
\texttt{alice} & \texttt{\{ AssistantProfessor \} } \\
\texttt{roy} & \texttt{\{ Professor \} } \\\bottomrule
\end{tabular}
\end{table*}

Therefore, { \texttt{\small IITPhdStudent} can be defined as:\\
\texttt{\small\{ Student, $\exists $isEnrolledIn.IITProgramme,\\ $\exists_{=1}$hasAdvisor.Professor,
$\Im$hasAdvisor.\\TeachingStaff, $\geq1$hasAdvisor.(TeachingStaff $\sqcap\,\neg$Professor)\}
}

Even though this approach works well for those concepts whose (axiomatized) definitions
contain only conjunctive clauses, it may generate incomplete descriptions when the definition contains a disjunctive clause.
For example, if the definition of  the concept \texttt{\small IITStudent} is of the form {\small\texttt{IITStudent $\equiv$ $\exists $isEnrolledIn.IITProgramme $\sqcap$(IITPhdStudent $\sqcup$ IIT\_MS\_Student)}}, the label-set of a newly introduced individual of \texttt{\small IITStudent} (say, \texttt{\small stud}) should be  {\small\texttt{\{IITPhdStudent $\sqcup$ IIT\_MS\_Student,\\ $\exists $isEnrolledIn.IITProgramme\}}}. However, our current label-set generation method will not include disjunctive clauses as such in the label-set, instead it will look for the satisfiability of each of the expression in  the disjunctive clause (that is, \texttt{\small IITPhdStudent(stud)} and \texttt{\small IIT\_MS\_Student(stud)}), and include them in the label-set, if they are true. But, for \texttt{\small stud}, they will not be true as we are not explicitly adding any other facts into the ontology other than \texttt{\small IITStudent(stud)}. Therefore, we will get the label-set as {\small\texttt{\{IITStudent, $\exists $isEnrolledIn.IITProgramme\}}} which is an incomplete label-set of the concept. On doing the next steps -- removing the concept name itself from the label-set, and doing a semantic-refinement over it -- the incompleteness persists. 
To overcome issue, after semantic refinement step, we will enrich the refined label-set with the previously encountered  disjunctive clause(s). That is, we get the new refined label-set of {\small \texttt{stud}} as {\small\texttt{\{IITPhdStudent $\sqcup$ IIT\_MS\_Student, $\exists $isEnrolledIn.IITProgramme\}}}.

\section{Natural Language Descriptions from the Refined Label-sets\label{nl}}
In this paper, prime focus is given for the generation of redundancy-free descriptions of ontology entities represented  in the form of logical expressions. Appropriate  NL sentence generation of these logical forms is yet to be fully explored. 
 However, for  the completeness of the paper, we present a simple method which we have adopted to generate NL descriptions of individuals and concepts from their refined label-sets. 

 NL description of an entity is defined as the set of NL fragments which describes the class names and role restrictions it satisfies. An example of a description of 
 \texttt{tom} is:
 \begin{equation}\centering
\framebox{
\begin{minipage}{7cm}
 \texttt{tom:} {is a student, enrolled in at least one IIT programme, and has exactly one professor as advisor}
 
 \end{minipage}
}\nonumber
\end{equation}

We consider a template similar to the following regular expression (abbreviated as regex) for generating  descriptions of individuals and concepts.
\begin{equation}
\begin{minipage}{8cm}
\texttt{Individual/concept:} (\enquote{is}) \big((\enquote{a}) ClassName \\(\enquote{,} $|$ \enquote{and})$?\big)^{+}$ \big( RoleRestriction (\enquote{,} $|$ \enquote{and})$?\big)^{+}$
\end{minipage}\nonumber
\end{equation}

\begin{table*}[!th]\centering\caption{Examples of the descriptions of individuals and concepts from  PD, HP and GEO ontologies, generated using the proposed and as well as the  traditional approaches \label{eg1}}
\begin{tabular}{@{}p{1cm}p{6 cm}p{7 cm}p{1.0cm}@{}}\toprule
\emph{Entity type}&\emph{Proposed approach}&\emph{Traditional approach}&\emph{Ontology}\\\midrule

Indivl.&\emph{Bird cherry Oat Aphid:} is a biotic-disorder, having  at least one pest  and all its  factors are pests.& \emph{Bird cherry Oat Aphid:} is a disorder, bio-disorder, pest damage and insect damage. It is all the following: has as factor only pest-insect, has as factor only pest, has as factor only organism and has as factor something.&PD\\\midrule

Indivl.&\emph{Black Chaff:} is a plant bacterioses, having at least one microorganism and all its factors are  microorganism. & \emph{Black Chaff:} is a disorder, a biotic disorder and a plant bacterioses. It is all the following: has as factor bacterioses, has as factor only organism, has as factor at least 1 thing, has as factor only micro-organism.&PD \\\midrule

Concept&\emph{Mite Damage:} is a biotic-disorder,  having at least one mite pest and all its factors are  mite pests. & \emph{Mite Damage:} is a disorder, a biotic-disorder and a pest damage. It is all the following:  has as factor only organism, has as factor only pest, has as factor only mite pest, has as factor at least 1 thing.&PD \\\midrule

Indivl.&\emph{Hermione Granger:} is a Hogwarts Student, a muggle, a gryffindor, having exactly one cat as pet.

&\emph{Hermione Granger:} is a Hogwarts student, a student, a human, a muggle, a gryffindor. It is all the following: has a pet, has as pet a cat, has as pet only creature, has at least 1 creature, has at most 1 creature, as pet.  &HP  \\\midrule

Concept&\emph{Hogwarts Student:} is a Student, is a Gryffindor or Hufflepuff or Ravenclaw or Slytherin, and having exactly one pet.

&\emph{Hogwarts Student:} is a student, a human, is a Gryffindor or Hufflepuff or Ravenclaw or Slytherin. It is all the following: has a pet, has as pet only creatures, has at least 1 creature, has at most 1 creature.  &HP  \\\midrule

Indivl.&\emph{Hedwig:} is an owl, is related to at least one Hogwarts student and only Hogwarts student, as pet.

&\emph{Hedwig:} is an owl, a pet, a creature. It is all the following: is pet of only Hogwarts student, is pet of a Hogwarts student.&HP \\\midrule

Indivl.&\emph{Jersey:} is a subnational entity, a government organization and  is related to exactly one sovereign state as a member.
&\emph{Jersey:} is a geopolitical dependency, an organization, a governmental organization, an Independent continuant, a subnational entity  and is a member of exactly one sovereign state.
&GEO\\\midrule

Indivl.&\emph{Florida:} is a government organization, is related to at least one nation as a part, and is related to exactly one sovereign state as a member.
&\emph{Florida:} is a major administrative subdivision, an organization, a governmental organization, an Independent continuant, a subnational entity. It is all the following: is a part of at least one nation, and is a member of exactly one sovereign state.
&GEO\\

\bottomrule
\end{tabular}
\end{table*}

In the above regex, ClassName specifies the concept names in the label-set. We use the \texttt{rdfs:label} role values of the class names as the ClassName. If \texttt{rdfs:label} role is not available, the local names of the URIs are used as the ClassName. For RoleRestriction, the role restrictions in the label-set are utilized. The role restrictions are treated in parts. We first tokenize the role names in the constraints. Tokenizing includes word-segmentation and processing of camel-case, underscores, spaces, punctuations etc. Then, we identify and tag the verbs\footnote{In the absence of a proper verb, the phrase \enquote{related to} is used in its place.} and nouns in the segmented phase --- as R-\texttt{verb}, R-\texttt{noun} respectively --- using the Natural Language Tool Kit\footnote{Python NLTK: http://www.nltk.org/}.  
We then incorporate these segmented words in a \emph{constraint-specific template}, to form a RoleRestriction. For instance, the restriction \emph{$\exists$hasAdvisor.Professor} is verbalized to \enquote{has at least 1 professor as advisor}, using the template: {\small$<\!\!\!\text{R-}\texttt{verb}\!\!\!>$ at least  $<\!\!\!n\!\!\!> <\!\!\!C\!\!\!>$ as $<\!\!\!\text{R-}\texttt{noun}\!\!\!>$} (where C corresponds to the concept present in the restriction). Constraint-specific templates corresponding to the possible restrictions in a label-set are listed in Table-\ref{t1}. In our studies, we have also tried out variants of these constraint-specific templates to further tune the NL output. Since the empirical study (see the next section) is done for a different intention, involving only a carefully chosen participants, we refrain from further enhancing the fluency of the NL texts.


If the $C$ equivalent portion of the restriction is not a concept name (atomic concept), that is, if it a conjunction or disjunction of restrictions, Table~\ref{t1} will be recursively looked up for possible templates, and the conjunctions and disjunctions will be replaced with \lq and\rq and   \lq or\rq~respectively. 

When it comes to  generating concept definitions, we can expect clauses containing disjunctions (independent expressions combined using disjunctions) in the refined label-set. They are handled in parts by taking each of those  independent expressions in the clause separately for NL generation, and, they are then  combined using \lq or\rq .

\section{Empirical Evaluation\label{eval}}
We present two case studies to explore the applicability of the redundancy-free description of individuals and concepts in validating the domain knowledge. Rather than choosing an ontology under development, we study the cases of validating two previously built ontologies. 

In the study, domain experts were presented with two representations of the same knowledge: one is by direct verbalization of the label-sets  and the other is by verbalizing them after finding the corresponding refined label-sets. Direct verbalization of a label-set generates texts (or descriptions) which are similar to those texts  which are produced by an existing ontology verbalizer --- we call this method as \emph{traditional approach}, and the other as the \emph{ proposed approach}. Examples for  the description texts that are generated using the proposed approach and traditional approach, from the Plant Disease (PD) ontology\footnote{http://wiki.plantontology.org/index.php/Plant\_Disease\_Ontology}, HarryPotter (HP) ontology\footnote{https://sites.google.com/site/ontoworks/ontologies} and Geographical Entity\footnote{https://bitbucket.org/uamsdbmi/geographical-entity-ontology/src (last accessed: 27/11/2015)} 
(GEO) ontologies are 
 given in Table~\ref{eg1}. One can clearly see that those descriptions which are generated using the proposed approach are compact, precise and easy-to-understand when compared to those which are generated using the traditional approach. 

\paragraph{Scope of the study.} We have done the empirical study mainly for two reasons. Firstly, for finding whether the process of semantic-refinement is helpful in generating useful texts for describing the ontology. 
For this purpose, the experts were asked to rate their degree of understanding of the knowledge  in the scale: (1) poor; (2) medium; (3) Good. 

Secondly, to measure the usefulness of the generated sentences (i.e., the descriptions of individuals and concepts)  in validating the domain knowledge,  domain experts were told to choose from the options: (1) Valid (2) Invalid (3) Don't know (4) Cannot be determined. Significance of these options is that, if a participant is choosing the 4th option, it is likely that she finds it difficult to reach a conclusion on the validity of the  sentence presented. In addition, feedbacks are collected from the experts to get suggestions on improving the system.

  \paragraph{Dataset used.} We used two ontologies for generating descriptions. The first ontology is  Plant-Disease  ontology (PD ontology) 
developed by International Center of Agricultural Research in the Dry Areas (ICARDA), and the second one is a synthetic ontology, Data structures and Algorithms (DSA) ontology, developed by ORG group\footnote{https://sites.google.com/site/ontoworks/home} at IIT Madras\footnote{https://www.iitm.ac.in/}. More details about these ontologies are available at our project website\footnote{https://sites.google.com/site/ontoworks/projects}. The current version of PD ontology has 546 individuals, 105 concepts and 15 object properties. The DSA ontology has 333 individuals, 23 concepts, 33 object properties and 21 datatype properties.

\paragraph{Experimental setup.} For each of the individuals and  concepts in the two ontologies we have generated corresponding NL descriptions from their node label-sets as well as from their refined label-sets, using an implemented prototype of the system. Since manual evaluation of all the generated descriptions is difficult, a selected number of descriptions were utilized for the study. The set of descriptions of individuals for the study were selected by grouping the entire descriptions based on their label-sets and randomly choosing one individual's description from each group. 
The set of descriptions of concepts were selected from those set of descriptions  (generated from refined label-sets) which are highly different from their counterparts that are generated from their non-refined label-sets. 
 From PD ontology,  31 descriptions of individuals and 10 descriptions of concepts have been considered for evaluation. Similarly, for DSA ontology, 14 descriptions of individuals  and 17 descriptions of concepts  were chosen for evaluation.  Then, experts of the two domains were asked to review the verbalized descriptions. Majority ratings of the sentences were considered for finding the statistics.

\paragraph{Expert selection.} Seven experts of plant disease areas and fourteen experts of data structures and algorithms were involved in the study. The seven experts of PD domain have either  a masters degree or a doctorate degree in the plant disease or agriculture related areas. The fourteen experts of DSA domain have successfully completed the advanced data structures and algorithms course offered at IIT Madras. 

\subsection{Results and Discussions}
Fig~\ref{pd1}-\ref{dsa2} show the statistics w.r.t. the ratings given by the domain experts. Based on these statistics, we have answered  the following two questions.

\subsubsection{How does the semantic refinement help in improving the understandability of the verbalized knowledge?} The degree of understanding of each of these descriptions to the  domain experts can be identified by looking at the ratings (i.e., poor, medium or good) which they had chosen during the empirical study. If there exists an ambiguity in the description (due to its verbatim fidelity to OWL statements), they are expected to choose poor or medium as the level of understanding.  To confine the reasons for ambiguity to the fidelity  to OWL constructs alone, possible (manual) grammatical error corrections had been done on the generated text --- as we were not using any sophisticated NL generation techniques. Grammatical errors such as  subject-verb agreement errors,  verb tense errors, verb form errors, singular/plural noun ending errors and  sentence structure errors had been corrected.

\begin{figure}[th!]
  \centering
  \includegraphics[width=.45\textwidth]{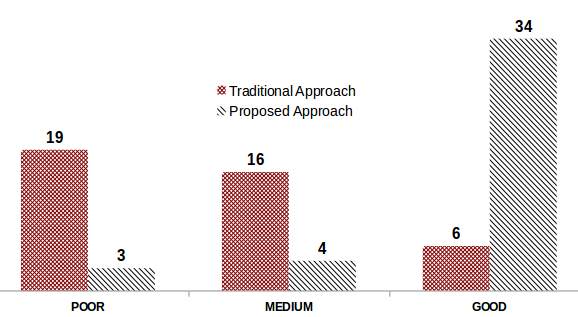}
  \caption{Y-axis shows the count of descriptions of a particular rating which are generated using our \emph{ proposed approach} and the \emph{traditional} approach  from the {\bf PD}  ontology\label{pd1}}
\end{figure}

\begin{figure}[th!]
  \centering
  \includegraphics[width=.45\textwidth]{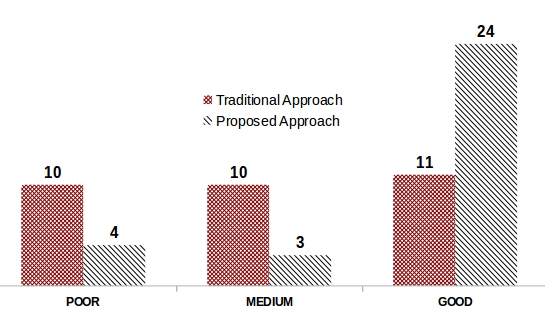}
  \caption{Y-axis shows the count of descriptions of a particular rating which are generated using our \emph{ proposed approach} and the \emph{traditional} approach   from the {\bf DSA} ontology\label{dsa1}}
\end{figure}

Fig~\ref{pd1} shows the overall responses which we received from the seven domain experts for the descriptions of PD ontology. We call it as the overall response because, ratings are calculated by looking at the majority responses; that is, only if a description is rated as  \lq good\rq~by at least 4 participants, it will be considered as a good description; similar is the case with poor and medium ratings. The dotted-bars represent the count of the descriptions of a particular rating which are generated using the proposed approach and the stripped-bars denote the count of those which are generated using the traditional approach. 
Similarly, Fig~\ref{dsa1} shows the statistics of the responses received for DSA ontology. 
For PD ontology, out of 41 descriptions which are generated using the proposed approach,  34 were rated as \lq{good}\rq, whereas for those which are generated using the traditional approach, only 6 out of 41 texts were rated as \lq{good}\rq. For DSA ontology, 24 out of  31 descriptions generated by proposed approach are \lq{good}\rq, only 11 descriptions that are generated using the traditional approach were rated as \lq{good}\rq.  These results highlight the significance of the semantic-refinement process in domain knowledge understanding. 

\begin{figure}[th!]
  \centering
  \includegraphics[width=.45\textwidth]{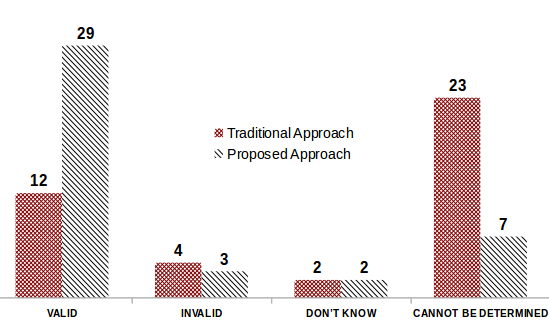}
  \caption{Statistics (based on the majority responses) to determine the usefulness of the generated descriptions in validating the {\bf PD} ontology \label{pd2}}
\end{figure}

\begin{figure}[th!]
  \centering
  \includegraphics[width=.45\textwidth]{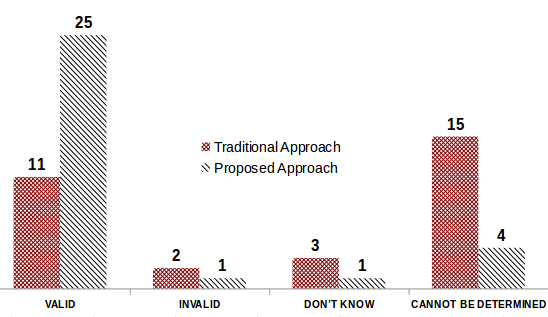}
  \caption{Statistics (based on the majority responses) to determine the usefulness of the generated descriptions in validating the {\bf DSA} ontology \label{dsa2}}
\end{figure}
\subsubsection{How does the semantic refinement helpful in knowledge validation?} 

Fig~\ref{pd2} and ~\ref{dsa2} show the statistics to determine the usefulness of the generated descriptions in validating the two domain ontologies, where,  as before, the dotted-bars represent the ratings of the descriptions that are generated from the proposed approach and the stripped-bars denote rating of the descriptions generated by the traditional approach.  
Usefulness of the generated descriptions in validating an ontology are obtained by looking at the number of descriptions which are marked as \lq{Cannot be determined}\rq. The three options: Valid, Invalid and Don't know, imply that the text is useful in getting into a conclusion, whereas the option \lq{Cannot be determined}\rq  indicates that there is some problem in the representation. 
From Fig~\ref{pd2} and Fig~\ref{dsa2}, in case of the proposed approach, only  7 out of 41 descriptions from PD ontology and 4 out of 31 descriptions from DSA ontology were not useful in determining the quality of the ontology, whereas in case of the traditional approach, approximately 50 percentage of the descriptions were not helpful. This clearly indicates that, verbalization after semantic-refinement is more effective in applications such as ontology validation.

\subsubsection{Discussion} 

The participants of our empirical study agree with the fact that, by reducing the redundancies in a description, the amount of time required for validating an individual description is reduced to a great extent.

Validation of an ontology also involves verifying the truthfulness of the property relationships in it, which is not addressed in this paper. This issue can be addressed in future by making use of the  edge-label-sets (label-sets for pairs of individuals -- see Section~\ref{els}), and mapping them to the respective constraint(s) in the node-label-set of the first individual. For e.g., $\text{\L}(a)=\{ C_{1},C_{2},\exists hasFriend.C_{3}\}$, and $\text{\L}(a,b)=\{hasFriend\}$, then $hasFriend$ in $\text{\L}(a,b)$ can be mapped to $\exists hasFriend.C_{3}$ in $\text{\L}(a)$. The description of $a$ can be generated as \enquote{$a$: is a $C_{1}$ and $C_{2}$, and has some $C_{3}$, like $b$, as Friend.}  Further investigation has to be done in this direction.

According to the domain experts, a persisting problem with any validation phase (especially when it involves  descriptions ontology entities and experts validating the verbalized knowledge) is that, when the ontology  becomes very large and complex, validation phase becomes a bottleneck for the entire development cycle. One way to overcome this issue is by considering only a relevant subset of individuals and concepts and their descriptions for validity check, so that, a rough estimate of the erroneous formalisms in the ontology can be identified quickly. Another direction of future work is a study on  the order at which the descriptions are to be presented to an expert so that an early detection of invalid knowledge can be made possible.

\section{Conclusion}
A novel method for verbalizing the definitions (called natural language descriptions)  of ontology entities  is presented in the paper. The descriptions are not merely verbatim translations of logical axioms of the ontology. Instead, they are generated from the set of logical restrictions satisfied by individuals and concepts of the ontology on which semantic simplification had been carried out. We propose a rule-based reduction approach for this purpose. We find that the proposed method indeed gives redundancy-free descriptions of individuals and concepts.

Our empirical studies based on two ontologies 
 have shown that the redundancy-free description of the domain knowledge is helpful in understanding the formalized knowledge more effectively and also useful in validating them.

 \section*{Acknowledgements}
 This project is funded by Ministry of Human Resource Development, Gov. of India. We express our fullest gratitude to the participants of our evaluation process: {Dr. S.Gnanasambadan} (Director of Plant Protection, Quarantine $\&$ Storage), Ministry of Agriculture, Gov. of India; 
 {Mr. J. Delince} and {Mr. J. M. Samraj}, 
Department of Social Sciences AC $\&$ RI, Killikulam, Tamil Nadu, India; {Ms. Deepthi.S} (Deputy Manager), Vegetable and Fruit Promotion Council Keralam (VFPCK), Kerala, India; {Dr.\, K.Sreekumar} (Professor) and students,  College of Agriculture, Vellayani, Trivandrum, Kerala, India.  We also thank all the undergraduate and post-graduate students of Indian Institute of Technology, Madras, who have participated in the empirical study. 
\bibliography{ref.bib}{}
\bibliographystyle{plain}

\section*{Appendix A}
%
%
%
%
%
%
%
%
%
%
%
%
%
%
\subsection*{Proofs for the rules in the rule-sets 5 to 7}
Here we  use proof-by-contradiction as the proof method. Given a rule of the form $P\equiv Q,$ we prove $P\sqsubseteq Q\sqcap Q\sqsubseteq P,$  by negating it and proving  $(P\sqcap\neg Q)\sqcup (Q\sqcap\neg P)$ as false.

Consider that  all the following rules are defined on an ontology \o~with $R$ and $S$ as its roles, and $U$ and $V$ are two of its concepts. 

\noindent\rule{7.6cm}{0.4pt}\vspace{.2cm}
\noindent Rule 5a: Given the ontology \o, $\exists R.U~\sqcap~\forall R.U \equiv \Im R.U$. The proof is trivial, and can be easily derived from the definition of  $\Im R.U$.

\noindent\rule{7.6cm}{0.4pt}\vspace{.2cm}

\noindent Rule 5b: If \o$\,\models\,U\!\sqsubseteq\!V, S\!\sqsubseteq\!R$, then for $\forall R.U~\sqcap~\exists S.V \,\equiv\,\Im R.U~\sqcap~\Im S.U.$

  Assume that 
$\forall R.U~\sqcap~\exists S.V~\sqcap~\neg( 
\Im R.U~\sqcap~\Im S.U
) $ is true. We can write it as: $\forall R.U~\sqcap~\exists S.V~\sqcap~\neg( 
(\exists R.U \sqcap \forall R.U) \sqcap (\exists S.U \sqcap \forall S.U)
) $ $\equiv$ $\forall R.U~\sqcap~\exists S.V~\sqcap~( 
\forall R.\neg U \sqcup \exists R.\neg U \sqcup \forall S.\neg U \sqcup \exists S.\neg U
) $ 
$\implies \forall R.U~\sqcap~ \forall S.U~\sqcap~\exists S.V$ $\sqcap~( 
\forall R.\neg U \sqcup \exists R.\neg U \sqcup \forall S.\neg U \sqcup \exists S.\neg U
) $ ~~~~(since $S\sqsubseteq R,  \forall R.U \implies  \forall S.U$)
$\equiv \forall R.U~\sqcap~ \forall S.U~\sqcap~\exists S.V$ $\sqcap~( 
\forall R.\neg U \sqcup \exists R.\neg U \sqcup \forall S.\neg U \sqcup \exists S.\neg U
) $ $\equiv 
(\underline{\forall R.U}$ $\sqcap~ \forall S.U~\sqcap~\exists S.V~\sqcap~\underline{\forall R.\neg U} ) \sqcup
  ( \underline{\forall R.U}~\sqcap~ \forall S.U~\sqcap~\exists S.V~\sqcap$ $\underline{\exists R.\neg U} )$ $\sqcup
 ( \forall R.U~\sqcap~\underline{\forall S.U}~\sqcap~\exists S.V~\sqcap~\underline{\forall S.\neg U} ) \sqcup
 ( \forall R.U~\sqcap~ \underline{\forall S.U}~\sqcap~\exists S.V~\sqcap~\underline{\exists S.\neg U} ),$ contradiction.

Now, assume that  $  (\exists R.U \sqcap \forall R.U \sqcap \exists S.U \sqcap \forall S.U) \sqcap \neg(\forall R.U~\sqcap~\exists S.V)  $ is true. 
$\equiv  \exists R.U \sqcap \forall R.U \sqcap \exists S.U \sqcap \forall S.U \sqcap (\exists R.\neg U~\sqcup~\forall S.\neg V)$
$\equiv (\exists R.U \sqcap \forall R.U \sqcap \exists S.U \sqcap \forall S.U \sqcap  \exists R.\neg U )~\sqcup~(\exists R.U \sqcap \forall R.U \sqcap \exists S.U \sqcap \forall S.U \sqcap  \forall S.\neg V  $) $\equiv( \exists R.U \sqcap\underline{ \forall R.U} \sqcap \exists S.U \sqcap \forall S.U \sqcap \underline{ \exists R.\neg U} )~\sqcup~( 
 \exists R.U \sqcap \forall R.U \sqcap \underline{\exists S.U} \sqcap \forall S.U \sqcap  \forall S.\neg V \sqcap \underline{\forall S.\neg U}   $), (Since, $U\sqsubseteq V,$)  contradiction.

\noindent\rule{7.6cm}{0.4pt}\vspace{.2cm}

\noindent Rule 5c: If \o$\,\models\,V\!\sqsubseteq\!U, S\!\sqsubseteq\!R$, then for $\forall R.U~\sqcap~\exists S.V \,\equiv\,\Im R.U~\sqcap~\exists S.V.$

Assume that $\forall R.U~\sqcap~\exists S.V~\sqcap~\neg( 
\Im R.U~\sqcap~\exists S.V
) $ is true. We can write it as:
$\equiv$ $\forall R.U~\sqcap~\exists S.V$ $\sqcap~\neg( 
\exists R.U \sqcap \forall R.U \sqcap \exists S.V ) $  (by the deftn. of $\Im R.U$)
$\equiv$ $\forall R.U~\sqcap~\exists S.V~\sqcap~( 
\forall R.\neg U \sqcup \exists R.\neg U \sqcup \forall S.\neg V ) $ 
$\equiv$ $\underline{\forall R.U}~\sqcap~\exists S.V~\sqcap~\underline{\forall R.\neg U} 
\sqcup 
\underline{\forall R.U}~\sqcap~\exists S.V~\sqcap~\underline{\exists R.\neg U} 
\sqcup 
\forall R.U~\sqcap~\underline{\exists S.V}~\sqcap~\underline{\forall S.\neg V.} $

Now, assume that  $  (\Im R.U \sqcap \exists S.V) \sqcap \neg(\forall R.U~\sqcap~\exists S.V)  $ is true. 
$\equiv \forall R.U \sqcap \exists R.U \sqcap \exists S.V \sqcap (\exists R.\neg U \sqcup \forall S.\neg V)$
$\equiv \underline{\forall R.U} \sqcap \exists R.U \sqcap \exists S.V \sqcap \underline{\exists R.\neg U} \sqcup 
 \forall R.U \sqcap \exists R.U \sqcap \underline{\exists S.V} \sqcap \underline{\forall S.\neg V.}$

\noindent\rule{7.6cm}{0.4pt}\vspace{.2cm}

\noindent Rule 6a: If \o$\,\models\,U\sqsubseteq\,V, R\sqsubseteq\,S$, then for $n\ge m,$ $\ge\!nR.U\,\sqcap\,\ge\!mS.V\,\equiv\,\ge\!nR.U.$

Assume that, ${\ge\!nR.U}\,\sqcap\,\ge\!mS.V\,\sqcap\,\neg(\ge\!nR.U)$ is true. We can write it as: $\underline{\ge\!nR.U}\,\sqcap\,\ge\!mS.V\,\sqcap\,\underline{\le\!(n-1)R.U},$ Contradiction.

Now assume that, ${\ge\!nR.U\sqcap \neg(\ge\!\!nR.U}\,\sqcap\,\ge\!mS.V)$ is true. We can write it as: ${\ge\!nR.U\sqcap (\le\!(n-1)R.U}$ $\sqcup\,\le\!(m-1)S.V)$ 
$\equiv (\underline{\ge\!nR.U}\sqcap \underline{\le\!\!(n-1)R.U})  \sqcup   (\underline{\ge\!nR.U}\sqcap\underline{\le\!(m-1)S.V}),$ contradiction. In the second conjunctive clause $\ge\!nR.U \implies \ge\!nS.V$ (since $U\sqsubseteq\,V \& R\sqsubseteq\,S$) , for $n\ge m,$ $\ge\!nR.U\,\sqcap$ $\le\!(m-1)S.V$ is a contradiction.

\noindent\rule{7.6cm}{0.4pt}\vspace{.2cm}

\noindent Rule 6b: If \o$\,\models\,V\!\sqsubseteq\!U, S\!\sqsubseteq\!R$, then for $n\ge 1,$ $\exists R.U\,\sqcap\,\ge\!nS.V\,\equiv\,\ge\!nS.V.$

Assume that, $n\ge 1,$ $\exists R.U\,\sqcap\,\ge\!nS.V\,\sqcap\neg(\ge\!nS.V)$ is true. We can write it as: $\exists R.U\,\sqcap\,\underline{\ge\!nS.V}\,\sqcap\underline{\le\!(n-1)S.V},$ contradiction.

Now, assume that $\ge\!nS.V\sqcap\neg(\exists R.U\,\sqcap\,\ge\!nS.V)$ is true. We can write it as: $\ge\!nS.V\sqcap(\forall R.\neg U\,\sqcup$ $\le\!(n-1)S.V)$
$\equiv (\underline{\ge\!nS.V}\sqcap \underline{\forall R.\neg U}) \sqcup (\underline{\ge\!nS.V}\sqcap \underline{\le\!(n-1)S.V}),$ contradiction. The contradiction in the first conjunctive expression is because: $\ge\!nS.V\implies \exists S.V \implies \exists R.U$ which contradicts with $\forall R.\neg U.$

\noindent\rule{7.6cm}{0.4pt}\vspace{.2cm}

\noindent Rule 6c: If \o$\,\models\,U\!\sqsubseteq\!V$, then for $n=1,$ $\exists R.U\,\sqcap\,\le\!nR.V\,\equiv\,\exists_{=1}R.U\sqcap \exists_{=1}R.V.$

Assume that, $\exists R.U\,\sqcap\,\le\!nR.V\,\sqcap\,\neg(\exists_{=1}R.U\sqcap \exists_{=1}R.V)$ is true.  We can write it as: $\exists R.U\,\sqcap\,\le\!nR.V\,\sqcap\,\neg(\exists R.U \sqcap$ $\le 1 R.U\sqcap \exists_{=1}R.V)$  $\equiv (\underline{\exists R.U}\,\sqcap $ $\le\!1R.V\,\sqcap\,\underline{\forall R.\neg U}) \sqcup$ $(\exists R.U\,\sqcap\,\underline{\le\!1R.V}\,\sqcap\underline{\ge 2 R.U})\sqcup(\exists R.U\,\sqcap\,\le\!1R.V\,\sqcap\neg\exists_{=1}R.V),$ contradiction. The second conjunctive clause is a contradiction because: $\le\!1R.V\implies\le\!1R.U$ (since $U\sqsubseteq V$), which contradicts with $\ge 2 R.U.$ In the third conjunctive expression, $\exists R.U \implies \exists R.V$, now, $\neg(\exists_{=1}R.V)\sqcap \exists R.V \implies$ $\ge 2 R.V,$ which contradicts with $\le 1R.V.$

Now, assume that $\exists_{=1}R.U\sqcap\exists_{=1}R.V\sqcap \neg(\exists R.U\,\sqcap\,\le\!1R.V)$ is true. We can write it as: $\exists_{=1}R.U\sqcap\exists_{=1}R.V\sqcap (\forall R.\neg U\,\sqcup\,$ $\ge\!2R.V)$ 
$\equiv (\underline{\exists_{=1}R.U}\sqcap \exists_{=1}R.V\sqcap\underline{\forall R.\neg U})\sqcup(\exists_{=1}R.U\sqcap\underline{\exists_{=1}R.V}\sqcap\underline{\ge\!2R.V}),$ contradiction. 

\noindent\rule{7.6cm}{0.4pt}\vspace{.2cm}

\noindent Rule 6d: If \o$\,\models\,U\!\sqsubseteq\!V, R\!\sqsubseteq\!S$, then for a whole number $n,$ $\ge nR.U\,\sqcap\,\le\!nS.V\,\equiv\,\exists_{=n}R.U\sqcap \exists_{=n}S.V.$

Assume that, $\ge nR.U\,\sqcap\,\le\!nS.V\,\sqcap\,\neg(\exists_{=n}R.U\sqcap \exists_{=n}S.V)$ is true. We can write it as: $\ge nR.U\,\sqcap\,\le\!nS.V\,\sqcap\,\neg(\le nR.U\sqcap\ge nR.U \sqcap \le nS.V\sqcap\ge nS.V)$ $\equiv~\ge nR.U\,\sqcap\,\le\!nS.V\,\sqcap\,(\ge (n+1)R.U\sqcup\le (n-1)R.U \sqcup \ge (n+1)S.V\sqcup\le (n-1)S.V)$
{$\equiv~(\underline{\ge nR.U}\,\sqcap\,\le\!nS.V\,\sqcap\,\underline{\ge (n+1)R.U)}$ $\sqcup~
                 (\underline{\ge nR.U}~\sqcap~\le\!nS.V~\sqcap~\underline{\le (n-1)R.U})~\sqcup~ \\
                 (\underline{\ge nR.U\,\sqcap\,\le\!nS.V}\,\sqcap\,\underline{\ge (n+1)S.V})\sqcup
                 (\underline{\ge nR.U}\,\sqcap\,\le\!nS.V\,\sqcap\,\underline{\le (n-1)S.V}),$} contradiction. In the third conjunctive expression, $\ge nR.U\,\sqcap\,\le\!nS.V \implies \exists_{=n}S.V,$ which contradicts with $\ge (n+1)S.U.$
                 
                 Now, assume that $\exists_{=n}R.U\sqcap \exists_{=n}S.V\,\sqcap\,\neg(\ge nR.U\,\sqcap\,\le\!nS.V)$ is true. We can write it as: $\le R.U \sqcap {\ge nR.U} \sqcap \le nS.V \sqcap nS.V \sqcap (\le (n-1)R.U$ $\sqcup {\ge (n+1)S.V})$  $\equiv (\le R.U \sqcap \underline{\ge nR.U} \sqcap {\le nS.V} \sqcap nS.V \sqcap \underline{\le (n-1)R.U})$ $\sqcup 
                (\le R.U \sqcap \ge nR.U \sqcap \underline{\le nS.V} \sqcap nS.V \sqcap\underline{\ge (n+1)S.V}),$ contradiction.

\noindent\rule{7.6cm}{0.4pt}\vspace{.2cm}

\noindent Rule 7a: If \o $\models U\sqsubseteq V, R\sqsubseteq S$, then $\exists R.U    \sqcap   \exists_{=1}S.V   \equiv   \exists_{=1}R.U   \sqcap  \exists_{=1}S.V.$

Assume that, $ \exists R.U    \sqcap   \exists_{=1}S.V   \sqcap \neg(\exists_{=1}R.U   \sqcap  \exists_{=1}S.V)$ is true. That is, $\exists R.U \sqcap \le 1 S.V \sqcap \ge 1 S.V \sqcap (\ge 2R.U \sqcup \le 0 R.U \sqcup \ge 2 S.V \sqcup \ge 0 S.V)$ 
$\equiv (\exists R.U    \sqcap   \underline{\exists_{=1}S.V}   \sqcap \underline{\ge 2R.U}) \sqcup$
$ (\underline{\exists R.U}    \sqcap   \exists_{=1}S.V   \sqcap \underline{\le 0 R.U})  \sqcup$
$ (\exists R.U    \sqcap   \underline{\exists_{=1}S.V}   \sqcap \underline{\ge 2 S.V}) \sqcup$
$ (\exists R.U    \sqcap   \underline{\exists_{=1}S.V}   \sqcap  \underline{\le 0 S.V}),$ Contradiction. The contradiction in the first clause is because: since $U\sqsubseteq V \&   R\sqsubseteq S; \ge 2R.U\implies \ge 2S.V$; $\ge 2S.V$ contradicts with $\exists_{=1}S.V.$

Now assume that $ \exists_{=1}R.U   \sqcap  \exists_{=1}S.V   \sqcap \neg(\exists R.U    \sqcap   \exists_{=1}S.V)$ is true. We can write it as: $ \exists_{=1}R.U   \sqcap  \exists_{=1}S.V   \sqcap (\forall R.\neg U    \sqcup   \neg(\exists_{=1}S.V))$ $\equiv  (\underline{\exists_{=1}R.U}   \sqcap  \exists_{=1}S.V   \sqcap \underline{\forall R.\neg U})    \sqcup   
 (\exists_{=1}R.U   \sqcap  \underline{\exists_{=1}S.V}   \sqcap \underline{\neg(\exists_{=1}S.V)}),$ Contradiction.

\noindent\rule{7.6cm}{0.4pt}\vspace{.2cm}

\noindent Rule 7b: If \o $ \models U\sqsubseteq V, R\sqsubseteq S$, then $\Im R.U    \sqcap   \exists_{=1}S.V   \equiv   \exists_{=1}R.U   \sqcap  \exists_{=1}S.V  \sqcap \Im R.U.$

Assume that,  $\Im R.U    \sqcap   \exists_{=1}S.V   \sqcap  \neg(\exists_{=1}R.U   \sqcap  \exists_{=1}S.V  \sqcap \Im R.U)$ is true.

$\equiv \exists R.U  \sqcap \forall R.U  \sqcap {\exists_{=1}S.V}\sqcap (\neg(\exists_{=1}R.U ) \sqcup \neg(\exists_{=1}S.V) \sqcup \neg(\Im R.U))$
$\equiv (\exists R.U  \sqcap \forall R.U  \sqcap\underline{\exists_{=1}S.V} \sqcap \underline{\neg(\exists_{=1}R.U )}) \sqcup 
(\exists R.U  \sqcap \forall R.U  \sqcap \underline{\exists_{=1}S.V} \sqcap \underline{\neg(\exists_{=1}S.V)}) \sqcup 
(\underline{\exists R.U  \sqcap \forall R.U}  \sqcap \exists_{=1}S.V \sqcap \underline{\neg(\Im R.U)})),$ Contradiction.
The contradiction in the first conjunctive clause is because: given $x\in \exists R.U \sqcap \neg(\exists_{=1}R.U )$, it implies $x\in\,\ge\!1R.U\implies x \in\,\ge\!1S.V$ (since $R\sqsubseteq S$ and $U\sqsubseteq V$) which contradicts with $\exists_{=1}S.V$. In the third conjunctive clause, $\neg(\Im R.U) \equiv \neg(\forall R.U \sqcap \exists R.U) \equiv \exists R.\neg U \sqcup \forall R.\neg U$, both these cases contradict with $\exists R.U  \sqcap \forall R.U$.

Now assume that, $\exists_{=1}R.U   \sqcap  \exists_{=1}S.V  \sqcap \Im R.U \sqcap \neg(\Im R.U    \sqcap   \exists_{=1}S.V)$ is true.

$\equiv\exists_{=1}R.U   \sqcap  \exists_{=1}S.V  \sqcap \Im R.U \sqcap \neg(\forall R.U    \sqcap \exists R.U \sqcap  \exists_{=1}S.V)$
$\equiv (\underline{\exists_{=1}R.U}   \sqcap  \exists_{=1}S.V  \sqcap \Im R.U \sqcap \underline{\exists R.\neg U}) \sqcup$
$ (\underline{\exists_{=1}R.U}   \sqcap  \exists_{=1}S.V  \sqcap \Im R.U \sqcap \underline{ \forall R.\neg U}) \sqcup$
$ (\exists_{=1}R.U   \sqcap  \underline{\exists_{=1}S.V}  \sqcap \Im R.U \sqcap  \underline{\neg(\exists_{=1}S.V})),$ Contradiction.

\noindent\rule{7.6cm}{0.4pt}\vspace{.2cm}

\noindent Rule 7c: If \o$\models U\sqsubseteq V$, then $\exists_{=n}R.U\,\sqcap\,\ge m R.V\,\equiv\,\exists_{=n}R.U\,\sqcap\,\ge (m-n) R.(V\sqcup\neg U)$  for $m\ge n$.

Assuming that $\exists_{=n}R.U\,\sqcap\,\ge m R.V\,\sqcap\,\neg (\exists_{=n}R.U\\\sqcap\,\ge (m-n) R.(V\sqcup\neg U))$ is true.

$\equiv \exists_{=n}R.U\,\sqcap\,\ge m R.V\,\sqcap\,\neg (\le nR.U\,\sqcap\,\ge nR.U\\\sqcap\,\ge (m-n) R.(V\sqcup\neg U))$

$\equiv \exists_{=n}R.U\,\sqcap\,\ge m R.V\,\sqcap\,(\ge (n+1)R.U\,\sqcup\,\le (n-1)R.U\,\sqcup\,\le (m-n-1) R.(V\sqcup\neg U))$

{\small$\equiv \underline{\exists_{=n}R.U}\,\sqcap\,\ge m R.V\,\sqcap\,\underline{\ge (n+1)R.U}~~\sqcup$}

{\small$\underline{\exists_{=n}R.U}\,\sqcap\,\ge m R.V\,\sqcap\,\underline{\le (n-1)R.U}~~\sqcup$}

{\small$\underline{\exists_{=n}R.U\,\sqcap\,\ge m R.V}\,\sqcap\,\le (m-n-1) R.(V\sqcup\neg U)$} The contradictions in the first two conjunctive clauses are trivial, in the third clause, $\exists_{=n}R.U\,\sqcap\,\ge m R.V$ implies $\ge (m-n) R.(V\sqcup\neg U)$ which contradicts with $\le (m-n-1) R.(V\sqcup\neg U)$.\vspace{.2cm}

Now, assume that, $\exists_{=n}R.U\,\sqcap\,\ge (m-n) R.(V\sqcup\neg U)~\sqcap~\neg(\exists_{=n}R.U\,\sqcap\,\ge m R.V)$ is true.

$\equiv \exists_{=n}R.U\,\sqcap\,\ge (m-n) R.(V\sqcup\neg U)~\sqcap~(\neg(\exists_{=n}R.U)\\\sqcup\,\le m-1 R.V)$

{\small$\equiv (\underline{\exists_{=n}R.U}\,\sqcap\,\ge (m-n) R.(V\sqcup\neg U)~\sqcap~\underline{\neg(\exists_{=n}R.U)})~\sqcup
(\underline{\exists_{=n}R.U\,\sqcap\,\ge (m-n) R.(V\sqcup\neg U)}~\sqcap~\underline{\le (m-1) R.V})$} 

In the second conjunctive clause,  contradiction can be found as follows: an $x \in\,\ge (m-n) R.(V\sqcup\neg U)$ implies $x$ has more than $m-n~R$ relations to $\neg U\sqcap V$, since $x\in\exists_{=n}R.U,$ we can say that $x$ has more than $m-n+n~R$ relations to $V,$ which can be written as $x\in \ge mR.V$. Clearly, this contradicts with $\le (m-1) R.V$.

%
%
%
%
%
%
%
%
%
\end{document}